\documentclass[pdflatex,sn-basic,Numbered,iicol]{sn-jnl}

\usepackage{amsmath,amsfonts,amssymb}
\usepackage{graphicx}
\usepackage{booktabs}
\usepackage{multirow}
\usepackage{array}
\usepackage{url}
\usepackage{xcolor}
\usepackage{booktabs}
\usepackage{float} 

\usepackage{tikz}
\usepackage{pgfplots}
\usepgfplotslibrary{groupplots,statistics}
\usetikzlibrary{positioning,arrows.meta}
\pgfplotsset{compat=1.18}
\pgfplotsset{
    qwenKR/.style={
        color=blue!80!black,
        solid,
        mark=*,
        mark size=1.8pt,
        line width=1.1pt
    },
    qwenEM/.style={
        color=red!80!black,
        solid,
        mark=square*,
        mark size=1.8pt,
        line width=1.1pt
    },
    qwenVF/.style={
        color=green!45!black,
        solid,
        mark=triangle*,
        mark size=2.1pt,
        line width=1.1pt
    },
    llamaKR/.style={
        color=blue!80!black,
        dashed,
        mark=o,
        mark size=1.8pt,
        mark options={solid,fill=white},
        line width=1.1pt
    },
    llamaEM/.style={
        color=red!80!black,
        dashed,
        mark=square,
        mark size=1.8pt,
        mark options={solid,fill=white},
        line width=1.1pt
    },
    llamaVF/.style={
        color=green!45!black,
        dashed,
        mark=triangle,
        mark size=2.1pt,
        mark options={solid,fill=white},
        line width=1.1pt
    }
}
\usepackage{tabularx}
\usepackage{placeins}
\usepackage{threeparttable}

\newcolumntype{L}[1]{>{\raggedright\arraybackslash}p{#1}}
\newcolumntype{C}[1]{>{\centering\arraybackslash}p{#1}}

\title[LLM-Driven Key--Value Extraction in Noisy Document Settings]
{From Pixels to Pairs: A Comprehensive Benchmark of LLM-Driven Key--Value Extraction in Noisy Document Settings}

\author[1]{\fnm{Zahra} \sur{Anvari}}\email{zahraanvari@westcliff.edu}

\affil[1]{\orgdiv{College of Technology \& Engineering}, \orgname{Westcliff University}, \orgaddress{\city{San Francisco}, \state{CA}, \country{USA}}}

\hypersetup{hypertexnames=false}

\begin{document}

\abstract{
Large language models (LLMs) have demonstrated strong capabilities in document key--value pair (KVP) extraction, yet controlled evaluations of their robustness to optical character recognition (OCR) output remain limited. This leaves an important gap in understanding their reliability within real-world OCR-to-LLM pipelines. Unlike end-to-end Vision-Language Models (VLMs) that conflate visual perception with semantic extraction, modular pipelines allow errors to be explicitly isolated and audited. To bridge this evaluation gap, we introduce a controlled, large-scale benchmarking framework designed to distinguish downstream LLM extraction behavior from upstream OCR degradation. This benchmark evaluates 136 experimental configurations and 17,688 deterministic document-level inference instances across five instruction-tuned open-weight LLMs (2B–8B parameters), three complementary datasets, and four text-quality conditions. This extensive matrix pairs a full zero-shot comparison and targeted one- to three-shot experiments with a confirmatory sensitivity analysis comprising 40 configurations across 20 frozen demonstration sets. By separating Key Recall (annotated-field recovery) from Exact Match and Value F1 (exact and partial value recovery, respectively), we reveal whether OCR degradation affects field identification and value reproduction differently across models. Crucially, our findings challenge three practical evaluation assumptions: (1) that clean-text performance is a reliable proxy for real-world robustness, (2) that model rankings remain consistent across annotation-derived Gold and OCR-derived text conditions, and (3) that providing few-shot demonstrations monotonically improves extraction accuracy. By exposing vulnerabilities such as model-ranking reversals and unstable few-shot gains, this benchmark provides a rigorous and reproducible framework for evaluating LLM reliability under noisy document conditions. To facilitate reproducible research, our entire benchmarking framework, dataset splits, and evaluation scripts are open-sourced.
}


\maketitle

\section{Introduction}
\label{sec:introduction}

Key-value pair (KVP) extraction is central to Document AI (DocAI) applications such as invoice processing, receipt understanding, and form digitization \citep{jaume2019funsd,huang2019icdar2019}. Instruction-tuned large language models (LLMs) offer a flexible, text-only alternative to task-specific extraction models \citep{llama32024,jiang2023mistral,anil2024gemma}, framing information extraction as structured generation without task-specific fine-tuning. However, controlled evaluations of their robustness to optical character recognition (OCR)-derived input remain limited. Recent surveys indicate that evaluations of LLM-based document understanding frequently use clean, natively digital, or annotation-derived textual inputs, leaving robustness to OCR-derived text comparatively underexplored~\citep{anvari2025sdu,hu2024large}.

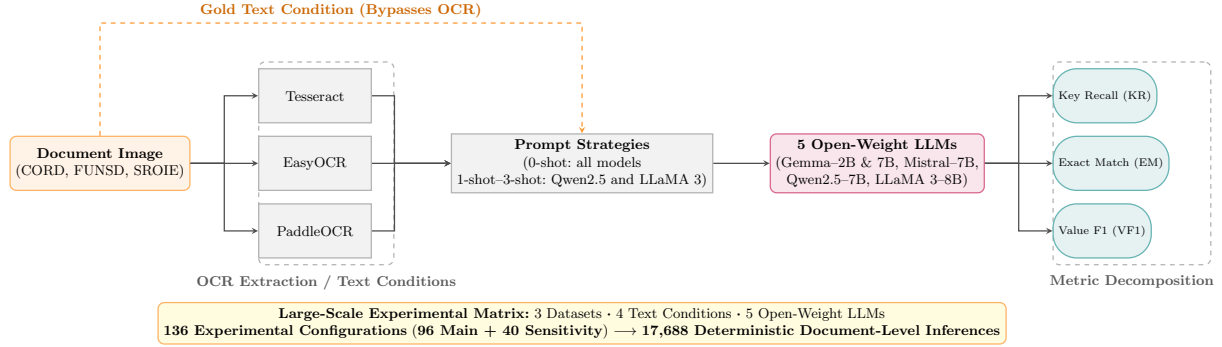
\begin{figure*}[htbp]
    \centering
    \resizebox{\textwidth}{!}{%
    \begin{tikzpicture}[
        node distance=1cm and 1cm, 
        box/.style={draw, rectangle, rounded corners, minimum height=1.2cm, minimum width=2.5cm, align=center, fill=blue!10, draw=blue!60, thick, font=\small},
        process/.style={draw, rectangle, minimum height=1.2cm, minimum width=2.5cm, align=center, fill=gray!10, draw=gray!60, thick, font=\small},
        eval/.style={draw, rectangle, rounded corners=15pt, minimum height=1.2cm, minimum width=2cm, align=center, fill=teal!10, draw=teal!60, thick, font=\footnotesize},
        arrow/.style={->, >=stealth, thick, draw=black!70},
        dashedbox/.style={draw, rectangle, dashed, rounded corners, minimum height=4.5cm, minimum width=3cm, align=center, draw=gray!60, thick},
        scalebanner/.style={draw, rectangle, rounded corners, minimum height=0.8cm, align=center, fill=yellow!15, draw=orange!60, thick, font=\small}
    ]
    
    \node[box, fill=orange!10, draw=orange!60] (doc) {\textbf{Document Image}\\(CORD, FUNSD, SROIE)};
    
    \node[dashedbox, right=of doc, xshift=0.5cm] (ocr_group) {};
    \node[below=0.1cm of ocr_group.south, font=\bfseries\small, text=black!65] {OCR Extraction / Text Conditions};
    
    \node[process, right=of doc, yshift=1.5cm, xshift=0.5cm] (tesseract) {Tesseract};
    \node[process, right=of doc, xshift=0.5cm] (easyocr) {EasyOCR};
    \node[process, right=of doc, yshift=-1.5cm, xshift=0.5cm] (paddleocr) {PaddleOCR};
    
    \node[process, right=of easyocr, xshift=0.75cm] (prompt) {\textbf{Prompt Strategies}\\(0-shot: all models\\ 1-shot--3-shot: Qwen2.5 and LLaMA 3)};
    
    \node[box, right=of prompt, xshift=0.25cm, fill=purple!10, draw=purple!60] (llm) {\textbf{5 Open-Weight LLMs}\\(Gemma--2B \& 7B, Mistral--7B,\\Qwen2.5--7B, LLaMA 3--8B)};
    
    \node[dashedbox, right=of llm, xshift=0.5cm, minimum width=3.5cm] (eval_group) {};
    \node[below=0.1cm of eval_group.south, font=\bfseries\small, text=black!65] {Metric Decomposition};
    
    \node[eval, right=of llm, yshift=1.5cm, xshift=0.5cm] (kr) {Key Recall (KR)};
    \node[eval, right=of llm, xshift=0.5cm] (em) {Exact Match (EM)};
    \node[eval, right=of llm, yshift=-1.5cm, xshift=0.5cm] (vf1) {Value F1 (VF1)};

    \draw[arrow] (doc.east) -- ++(0.75cm,0) |- (tesseract.west);
    \draw[arrow] (doc.east) -- (easyocr.west);
    \draw[arrow] (doc.east) -- ++(0.75cm,0) |- (paddleocr.west);
    
    \draw[arrow] (tesseract.east) -- ++(0.5cm,0) |- (prompt.west);
    \draw[arrow] (easyocr.east) -- (prompt.west);
    \draw[arrow] (paddleocr.east) -- ++(0.5cm,0) |- (prompt.west);
    
    \draw[arrow] (prompt.east) -- (llm.west);
    
    \draw[arrow] (llm.east) -- ++(0.75cm,0) |- (kr.west);
    \draw[arrow] (llm.east) -- (em.west);
    \draw[arrow] (llm.east) -- ++(0.75cm,0) |- (vf1.west);

    \draw[->, >=stealth, dashed, draw=orange!80, thick] (doc.north) |- ++(0, 2.5cm) -| node[pos=0.25, above, font=\bfseries\small, text=orange!80!black] {Gold Text Condition (Bypasses OCR)} (prompt.north);

    \node[scalebanner] (scale)
at ([yshift=-3.55cm]prompt.center)
{
\textbf{Large-Scale Experimental Matrix:}
3 Datasets $\boldsymbol{\cdot}$ 4 Text Conditions
$\boldsymbol{\cdot}$ 5 Open-Weight LLMs\\
\textbf{136 Experimental Configurations}
(\textbf{96 Main + 40 Sensitivity})
$\longrightarrow$
\textbf{17,688 Deterministic Document-Level Inferences}
};


    \end{tikzpicture}%
    } 
    \caption{The end-to-end benchmark architecture pipeline. A document image is processed through branching OCR engines (or a Gold Text bypass) to construct a prompt. The resulting text is fed into the LLMs, and the predictions are evaluated using three complementary metrics, Key Recall (KR), Exact Match (EM), and Value F1 (VF1), to distinguish annotated-field recovery from exact and partial value recovery. The bottom banner highlights the combinatorial scale of the executed experimental matrix.}
    \label{fig:pipeline_architecture}
\end{figure*}

Addressing this evaluation gap is critical for practical deployment. While recent advancements in Vision-Language Models (VLMs) such as GPT-4, LLaVA, and mPLUG-DocOwl \citep{openai2023gpt4,liu2024visual,hu2023mplug} offer end-to-end document understanding capabilities, decoupled OCR-to-LLM pipelines remain practically relevant in enterprise systems handling sensitive data, such as financial invoices or medical records. These modular pipelines remain attractive for two distinct reasons: they permit the independent selection of OCR and extraction components, and they allow the downstream semantic reasoning to be handled by smaller instruction-tuned models (such as the 2B–8B models evaluated here) that run on a single GPU. Because end-to-end VLMs jointly perform visual recognition, OCR-like transcription, layout reasoning, and semantic extraction within a single opaque architecture, they obscure the exact origin of failures. Including them in the same comparison would therefore conflate upstream perception with downstream extraction, preventing performance changes from being attributed specifically to textual fidelity, the central experimental variable in this study. Instead, we adopt a controlled component-level design that standardizes downstream inference and evaluation while varying the input text condition. 


In practice, document degradations such as blur, noise, shadows, and poor illumination severely reduce OCR reliability \citep{anvari2021survey}. Recognition errors, token fragmentation, numeric corruption, and altered reading order flatten a two-dimensional document into noisy text, removing spatial cues and disrupting the LLM's ability to associate keys with values. We therefore organize the study around three evaluation assumptions:

\begin{enumerate}
    \item \textbf{Clean-Text Robustness Proxy:} Models achieving high accuracy on clean text are proportionally robust to upstream OCR noise.
    \item \textbf{Rank Stability:} A model that outperforms its peers on digital documents will maintain its superiority on noisy, scanned documents.
    \item \textbf{The Few-Shot Guarantee:} Providing in-context demonstrations consistently and monotonically improves extraction accuracy in noisy environments.
\end{enumerate}

To empirically investigate these assumptions, we introduce a systematic benchmarking framework designed to distinguish downstream LLM extraction behavior from upstream OCR degradation. Because LLM performance is highly sensitive to prompt configuration, OCR engine, and output parsing, drawing robust conclusions requires evaluating models across a broad permutation of conditions. We achieve this by evaluating Form Understanding in Noisy Scanned Documents (FUNSD) \citep{jaume2019funsd}, Scanned Receipt OCR and Information Extraction (SROIE) \citep{huang2019icdar2019}, and CORD \citep{park2019cord} datasets using annotation-derived (Gold) text and outputs from Tesseract \citep{smith2007tesseract}, EasyOCR \citep{easyocr2020}, and PaddleOCR \citep{du2020paddleocr}. 
Our large-scale experimental matrix shown in Figure~\ref{fig:pipeline_architecture} executes 136 configurations across five open-weight models from the Gemma \citep{anil2024gemma}, Mistral \citep{jiang2023mistral}, Qwen2.5 \citep{qwen2024}, and LLaMA~3 \citep{llama32024} families. It combines zero-shot evaluation of all five models, targeted one- to three-shot evaluation of Qwen2.5--7B and LLaMA 3--8B, and 40 confirmatory sensitivity configurations, yielding 17,688 deterministic document-level inference instances. Our evaluation uses complementary metrics rather than a single aggregate score. \textbf{Key Recall} measures annotated-field recovery, while \textbf{Exact Match (EM) and Value~F1 (VF1)} measure exact and partial value recovery. This decomposition reveals whether OCR degradation affects field identification and value reproduction differently, without interpreting the metrics as direct measures of hallucination frequency. In summary, our primary contributions challenge the preceding assumptions and provide a rigorous, reproducible framework for evaluating in-context learning under noisy document conditions:

\begin{itemize}
    \item \textbf{A Large-Scale Evaluation Framework:} We present a comprehensive controlled benchmark comprising 136 configurations and 17,688 deterministic document-level inference instances across multiple document types, text conditions, models, and prompting settings.
    \item \textbf{Granular Metric Decomposition:} We separate annotated-field recovery, measured by Key Recall, from exact and partial value recovery, measured by EM and VF1, enabling granular analysis of how OCR degradation affects different aspects of extraction.
    \item \textbf{Exposing the Clean-Text Fallacy (Refuting Assumptions 1 \& 2):} We demonstrate that clean-text performance does not uniformly predict OCR-derived performance: relative to the corresponding Gold condition, the top-ranked zero-shot model changes in 10 of 27 dataset–OCR–metric comparisons, as detailed in Section~\ref{sec:results-cross}.
    \item \textbf{Unstable Few-Shot Dynamics (Refuting Assumption 3):} We reveal that in-context demonstrations do not guarantee monotonic accuracy improvements under OCR noise, exhibiting high sensitivity to demonstration complexity and composition.
    \item \textbf{Open-Source Reproducibility:} We release our entire benchmarking suite, including all configurations, dataset splits, and evaluation tools, providing the community with a framework to evaluate future LLMs. Code is available at \url{https://github.com/zanvari/pixels-to-pairs}.
\end{itemize}

\section{Related Work}
\label{sec:related_work}

\subsection{Document Key-Value Pair Extraction and Multimodal Models}
Early document extraction systems relied on templates, regular expressions, and
spatial heuristics \citep{shilman2005structured,ha2000formtemplatematching}.
Deep learning introduced representations that jointly model text and layout,
including Chargrid \citep{katti2018chargrid} and the FUNSD benchmark
\citep{jaume2019funsd}. Transformer-based models subsequently integrated
textual, spatial, and visual information: LayoutLM and its successors
\citep{xu2020layoutlm,xu2021layoutlmv2,huang2022layoutlmv3}, DocFormer
\citep{appalaraju2021docformer}, FormNet \citep{lee2021formnet}, StructuralLM
\citep{li2021structurallm}, and UDOP \citep{tang2022udop} established strong
supervised document-understanding approaches. Generative models such as Donut
\citep{kim2022donut}, DocVLM \citep{liu2024docvlm}, and LayoutLLM
\citep{huang2023layoutllm} incorporate document images more directly. These methods demonstrate the value of visual and spatial information but differ from our OCR-centered text-only setting, which isolates language-model behavior after document content has been converted to text. A broader review of multimodal document extraction trends is provided in \citep{anvari2025sdu}.

\subsection{LLMs for Structured Information Extraction}
Instruction-tuned LLMs can infer structured relationships from text without
task-specific fine-tuning. Mistral \citep{jiang2023mistral}, Gemma
\citep{anil2024gemma}, Qwen2.5 \citep{qwen2024}, and LLaMA~3 \citep{llama32024} provide zero-shot and few-shot
capabilities, while instruction tuning \citep{wei2022finetuned} and frameworks
such as InstructUIE \citep{wang2023instructuie} demonstrate unified
prompt-based interfaces for information extraction. Comparisons remain
difficult, however, because studies vary in modality, supervision, prompting,
decoding, and post-processing. The interaction between OCR transcription
quality and prompt-based text-only KVP extraction is particularly
under-characterized across multiple OCR engines, LLM families, and document
structures.

\subsection{OCR Noise and Downstream Text Degradation}
Document degradations can produce character substitutions, token fragmentation,
line reordering, and numeric corruption during OCR \citep{anvari2021survey},
with direct consequences for KVPs containing dates, monetary values,
identifiers, and addresses. We evaluate outputs from Tesseract
\citep{smith2007tesseract}, EasyOCR \citep{easyocr2020}, and PaddleOCR
\citep{du2020paddleocr} under matched annotation-derived and OCR-derived
conditions. By holding downstream inference and evaluation fixed across three
document benchmarks and zero- to three-shot prompting, the study isolates how
upstream textual fidelity interacts with document structure, model behavior,
and demonstration selection.



\section{Datasets}
\label{sec:datasets}

\begin{figure}[H]
\centering
\setlength{\fboxsep}{0.5pt}
\setlength{\tabcolsep}{1pt}

\begin{tabularx}{\columnwidth}{
@{}>{\centering\arraybackslash}X
>{\centering\arraybackslash}X
>{\centering\arraybackslash}X@{}
}
\fbox{\includegraphics[
    height=2.55cm,
    keepaspectratio
]{funsd_example.png}}
&
\fbox{\includegraphics[
    height=2.55cm,
    keepaspectratio
]{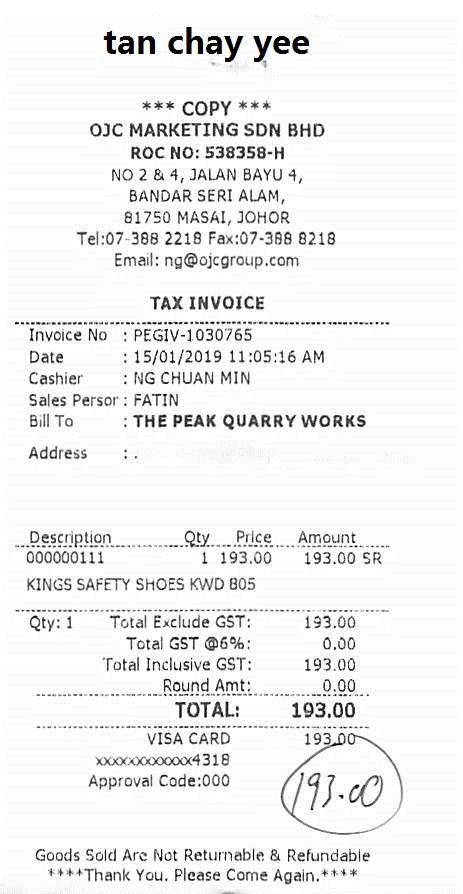}}
&
\fbox{\includegraphics[
    height=2.55cm,
    keepaspectratio
]{cord_example.png}}
\\[-0.5mm]
{\scriptsize (a) FUNSD}
&
{\scriptsize (b) SROIE}
&
{\scriptsize (c) CORD}
\end{tabularx}

\caption{Example documents from the evaluated test splits:
(a) FUNSD, (b) SROIE, and (c) CORD.}
\label{fig:dataset_examples}
\vspace{-2mm}
\end{figure}

We evaluate three document-understanding benchmarks: FUNSD forms, SROIE receipts, and CORD receipts (Fig.~\ref{fig:dataset_examples}). They differ in layout dependence, annotation schema, textual density, and numeric content, enabling comparison across structurally distinct extraction settings. 

\subsection{CORD}
CORD \citep{park2019cord} contains retail receipts with variable
length, layout, and item density. Numerous line items and repeated monetary
values make key--value alignment challenging, while OCR digit corruption can
directly affect value recovery. CORD therefore represents a dense,
numerically intensive receipt setting.

\subsection{FUNSD}
FUNSD \citep{jaume2019funsd} contains scanned forms whose key--value
relationships often depend on two-dimensional spatial organization. Converting
these forms to plain text removes layout cues and can obscure associations
between keys and distant values, making FUNSD particularly challenging for a
text-only extraction pipeline.

\subsection{SROIE}
SROIE \citep{huang2019icdar2019} contains relatively short receipts annotated
with four canonical fields: \emph{company}, \emph{date}, \emph{address}, and
\emph{total}. Its compact schema reduces structural ambiguity, but OCR changes
in addresses and numeric values can strongly affect exact and partial value
recovery. Because only the annotated field inventory is scored, additional
semantically meaningful receipt fields are outside the primary evaluation.

Together, FUNSD emphasizes layout-dependent associations, SROIE evaluates a
compact receipt schema, and CORD stresses extraction from longer, numerically
dense receipts. Dataset-specific construction of the annotation-derived text
used in the benchmark is described in Section~\ref{sec:methods}.

\section{Methodology}
\label{sec:methods}

This section defines the KVP extraction task and the controlled experimental
protocol used across models, datasets, prompting regimes, and text conditions.

\subsection{Task Formulation}

Given a document image $I$, a text representation $T$ is obtained either from
dataset annotations or from an OCR engine. The objective is to recover

\[
\mathcal{P}=\{(k_i,v_i)\}_{i=1}^{N},
\]

where $k_i$ is a semantic key (e.g., \texttt{TOTAL}, \texttt{DATE}) and
$v_i$ its associated value. OCR may introduce character substitutions, token
fragmentation, numeric corruption, and reading-order errors; flattening forms
into text can additionally remove spatial cues needed to associate keys and
values.

\subsection{Models Evaluated}

The benchmark evaluates five instruction-tuned, open-weight decoder-only checkpoints without task-specific fine-tuning: Gemma-2B-it and Gemma-7B-it~\citep{anil2024gemma}, Mistral-7B-Instruct-v0.2~\citep{jiang2023mistral}, Qwen2.5-7B-Instruct~\citep{qwen2024}, and LLaMA-3-8B-Instruct ~\citep{llama32024}. These checkpoints were selected to provide coverage across four widely used model families and the practical 2B--8B parameter range while remaining feasible for single-GPU academic experimentation.


\subsection{Prompting Strategy}

All models use a unified prompt that defines the extraction task, specifies the required JSON output structure, and discourages unsupported fields. The overall template is shared across datasets, with only schema-dependent wording changed;
we do not perform model-specific prompt optimization.

All models are evaluated zero-shot. Qwen2.5--7B and LLaMA~3--8B are also
evaluated with one, two, and three demonstrations drawn exclusively from the
corresponding training split and disjoint from evaluation documents.

For FUNSD, demonstration inputs use the same spatially reconstructed
annotation text as the Gold text condition (Section~\ref{sec:text-input}), and targets are generated directly from annotated question--answer links. Selected demonstration texts and targets were verified against the training annotations.
The same clean training demonstrations are held fixed across the evaluated Gold and PaddleOCR few-shot conditions, so the text source varies only for the target document.

All datasets share the same JSON output contract, but the extraction scope follows the benchmark annotation structure. The SROIE prompt explicitly enumerates its four canonical fields i.e. company, address, date, and total, whereas the FUNSD and CORD prompts request all reasonable document-grounded KVPs because their evaluated field inventories are document dependent.

\subsection{Output Parsing and Canonicalization}
\label{sec:output-parsing}

Outputs are parsed deterministically and without semantic correction. We first parse the complete response as the requested JSON object. If this fails, a conservative recovery step retains only complete, independently decodable dictionary objects with the expected structure. Incomplete objects are not repaired, and no regex extraction, document-text fallback, ground-truth-aware recovery, or semantic heuristics are used. Unparseable responses are treated as empty predictions.

Formatting-tolerant key normalization is applied only during scoring (Section~\ref{sec:evaluation}); it does not use fuzzy matching, stemming, synonym expansion, or semantic key mapping.

\subsection{Datasets and Evaluation Splits}

The benchmark uses FUNSD forms, CORD receipts, and SROIE receipts. Few-shot demonstrations come only from the corresponding training data and are disjoint from evaluation documents. FUNSD contains 50 original test documents; three contained no evaluable ground-truth KVPs under the benchmark preprocessing and evaluation protocol, leaving 47 evaluable documents. The same 47-document subset is used across all FUNSD models, OCR conditions, and prompting configurations. CORD and SROIE contain 100 and 347 evaluated documents, respectively. The excluded documents from FUNSD test split were 85201976, 85540866, and 85629964; each yielded zero annotated ground-truth KVPs after extracting valid question–answer links, making KR, EM, and VF1 undefined.

\subsection{Text Input Construction and Conditions}
\label{sec:text-input}

Each model is evaluated under a clean annotation-derived text condition and
three OCR-derived conditions. Because the datasets provide different annotation
structures, clean text is constructed using a dataset-specific but
deterministic procedure.

\subsubsection{FUNSD}
FUNSD annotations provide text blocks and bounding boxes $[x_0,y_0,x_1,y_1]$. After removing empty blocks and invalid boxes, we group blocks into lines when their vertical overlap is at least $0.50$ or their vertical-center distance is at most $0.40$ times the maximum of the block height and the candidate line's median block height. Multiple qualifying lines are resolved using overlap and normalized center distance; blocks are sorted left to right within lines and lines top to bottom
by median center, with minimum $x_0$ breaking ties. Spaces join blocks and newlines separate lines. These rules were fixed
before the corrected downstream runs and were not tuned against KVP performance. We call the result \emph{spatially reconstructed Gold text}: a deterministic approximation rather than a uniquely canonical reading order.

\subsubsection{CORD}
CORD clean text is reconstructed from its structured annotations by traversing
the annotated valid lines and their constituent words in the order provided by
the dataset and concatenating the annotated word text. We audited the resulting
clean-text files against this reconstruction procedure and obtained exact
agreement for all 100 evaluated documents.

\subsubsection{SROIE}
SROIE clean text is constructed from the dataset's provided OCR
box--transcription annotations while preserving the annotated document-text
ordering. An audit against an independent spatial reconstruction showed
equivalent token content for 346 of 347 documents. The sole discrepancy was an
encoding/replacement-character artifact in the source annotation; the existing
clean transcription was retained rather than introducing the corrupted
character into the reference text.

\subsubsection{OCR-Derived Text}
For all engines, the LLM received the engine's native text sequence; bounding boxes were neither included nor geometrically re-sorted. Tesseract output came directly from \texttt{pytesseract.image\_to\_string} with English recognition. EasyOCR \texttt{readtext(..., detail=0)} strings and PaddleOCR 3.x \texttt{rec\_texts} were joined with newline characters in API order; PaddleOCR orientation classification, document unwarping, and text-line orientation were disabled. Thus, the OCR inputs retain engine-specific recognition, tokenization, line-grouping, and reading-order effects and represent a native plain-text rather than geometry-aware OCR-to-LLM pipeline.

\subsection{Implementation and Inference Settings}

All experiments use deterministic greedy decoding
(\texttt{do\_sample=False}), standard model chat templates when available, and
no model-specific prompt tuning, reranking, semantic post-processing, or
decoding modifications. The unified Hugging Face \texttt{transformers}
pipeline processes documents independently and stores raw generations, parsed
predictions, per-document scores, prompt lengths, truncation flags, generated
token counts, and output-limit events. Fixed token budgets, software versions,
and hardware are reported in Section~\ref{sec:experimental-setup}.

\subsection{Confirmatory Sensitivity Analysis}
\label{sec:demo-sensitivity-method}

Because the primary few-shot experiments showed non-monotonic responses to
additional demonstrations, we conducted a confirmatory analysis of
\emph{demonstration composition}. It uses FUNSD spatially reconstructed Gold
text with Qwen2.5--7B and LLaMA~3--8B and fixes the demonstration count at
three, so only composition varies.

\subsubsection{Candidate Construction}
Candidates came exclusively from the FUNSD training split and required
(i) nonempty corrected spatial-Gold text, (ii) at least one linked
question--answer pair, and (iii) nonempty linked keys and values. This yielded
106 eligible documents. Eight documents used in earlier pilot sets were
excluded to keep the confirmatory analysis independent, leaving 98 candidates;
no test document was used for selection.

\subsubsection{Structural Complexity Profile}
Each eligible training document was characterized using eight predefined
training-only structural features:
(1) number of annotation nodes,
(2) number of linked KVPs,
(3) number of unique normalized keys,
(4) number of keys occurring more than once,
(5) number of repeated-key occurrences beyond the first,
(6) mean value length in words,
(7) maximum value length in words, and
(8) number of words in the spatially reconstructed Gold text.

Each feature was converted to its percentile rank over eligible training
documents, and document complexity was defined as the mean percentile rank:

\begin{equation}
C(d)=\frac{1}{8}\sum_{j=1}^{8}R_j(d),
\end{equation}

where $R_j(d)$ is the percentile rank for feature $j$. Thus all eight
dimensions contribute equally, and the score uses only training-document
structure and text statistics---never model predictions or test performance.

The 25th and 75th percentiles of the resulting training complexity
distribution were $Q_{25}=0.3578$ and $Q_{75}=0.6557$, respectively. These
thresholds were computed from the eligible training-document complexity
distribution, before any model inference or inspection of test performance.
After the pilot-document exclusions, the candidate pool contained 26
low-complexity documents ($C(d)\leq Q_{25}$), 46 representative documents
($Q_{25}<C(d)<Q_{75}$), and 26 high-complexity documents
($C(d)\geq Q_{75}$). The terms \emph{Low}, \emph{Representative}, and
\emph{High} therefore denote algorithmically defined strata of this
training-only complexity distribution; they are not human-assigned difficulty
labels and do not encode whether an LLM succeeds or fails on a document.

\subsubsection{Frozen Demonstration Sets}
We constructed four demonstration-selection strategies:
\emph{Low}, containing three low-complexity examples;
\emph{Representative}, containing three examples from the middle complexity
range;
\emph{High}, containing three high-complexity examples; and
\emph{Diverse}, containing one Low, one Representative, and one High example.

Five independent three-document sets were generated per strategy (20 sets
total). Selection was fully algorithmic and reproducible. Within each stratum,
candidate identifiers were first sorted by document identifier and then
permuted using sequential draws from a single NumPy random-number generator
initialized with fixed seed 20260915. The first 15 permuted documents in each
stratum were partitioned consecutively into five homogeneous three-shot sets.
The next five documents from each of the Low, Representative, and High strata
were combined position-wise to form five Diverse sets, each containing one
document from each stratum.

No document was reused across the 20 confirmatory sets, yielding 60 unique
demonstrations. No human annotator assigned documents to complexity strata or
selected examples based on perceived difficulty, representativeness, model
behavior, or expected usefulness as demonstrations. Manual inspection was
limited to a source-integrity audit verifying that the already selected
documents had valid annotation and spatial-Gold sources, nonempty text, and
valid linked KVPs; this audit did not alter the complexity labels or use model
outputs. The complete set manifest, including document identifiers, strata,
complexity definition, quartile thresholds, and random seed, was frozen before
confirmatory inference. No set or document was subsequently selected,
replaced, reordered, or modified on the basis of downstream KR, EM, VF1, model
outputs, or test-set behavior.

\subsubsection{Confirmatory Evaluation}
The 20 frozen sets were evaluated independently with both models, yielding 40 configurations and 1,880 document-level inference instances over the
same 47 FUNSD test documents. Prompt structure, corrected spatial-Gold input,
decoding, and evaluation match the main benchmark. Results are summarized
across valid sampled sets within each strategy rather than selecting the
best-performing set.

\subsection{Experimental Matrix and Computational Setup}
\label{sec:experimental-setup}

For each dataset, five models are evaluated zero-shot under Gold, PaddleOCR,
EasyOCR, and Tesseract inputs (20 configurations). Qwen2.5--7B and
LLaMA~3--8B additionally use one-, two-, and three-shot prompting under Gold
and PaddleOCR (12 configurations), giving 32 configurations per dataset and 96
overall. With 47 FUNSD, 347 SROIE, and 100 CORD documents, the main matrix
comprises 15,808 document-level inference instances. Adding the 40 confirmatory FUNSD configurations and their 1,880 instances yields 136 experimental configurations and 17,688 document-level inference instances.

All final runners use an 8,192-token input budget, at most 1,024 generated
tokens, batch size one, deterministic greedy decoding
(\texttt{do\_sample=False}), standard chat templates when available, and fixed
NumPy/PyTorch seeds. Generation parameters are not tuned separately by model.

The final environment used Python~3.13.15, PyTorch~2.11.0 with CUDA~12.8,
Transformers~5.16.1, and an NVIDIA A100 GPU with 40~GB memory. Models were run
sequentially and document-level artifacts persisted for resumable execution.
Cache cleanup between model blocks, when required, did not alter model weights,
prompts, generation settings, or saved outputs.








\section{Evaluation Methodology}
\label{sec:evaluation}

Predictions are evaluated with the same deterministic normalization, matching,
and scoring procedure under every text condition. The models receive only the
supplied text---no image, bounding boxes, or layout coordinates.


\subsection{Evaluation Scope}

For ground truth
\[
\mathcal{P}=\{(k_i,v_i)\}_{i=1}^{N}
\]
and model prediction $\hat{\mathcal{P}}$, evaluation is reference-centered:
Key Recall (KR), Exact Match (EM), and Value~F1 (VF1) quantify recovery of the
$N$ annotated KVPs. Predictions whose normalized keys do not correspond to an
annotated key do not contribute to these metrics; additional fields are
considered descriptively during output inspection. Thus, the primary metrics
measure annotated-information recovery rather than precision over every field
generated by the model.

\subsection{Normalization and Matching}

Keys are normalized deterministically using Unicode NFKC normalization,
case-folding, decorative-punctuation normalization, separator-to-whitespace
mapping, and repeated-whitespace collapse. This removes superficial formatting
variation but does \emph{not} use fuzzy matching, stemming, synonym expansion,
semantic similarity, or dataset-specific aliases. Values use normalized text
and whitespace tokenization for token-level scoring.

Keys match by exact equality after normalization. Duplicate occurrences are
preserved and predictions cannot be reused. For each ground-truth pair, Value
F1 is computed against unused predictions with the same normalized key; when
multiple candidates exist, the one maximizing token-level F1 is selected and
marked used. If none exists, the pair receives Value~F1 zero.

\subsection{Evaluation Metrics}

We report three complementary metrics: Key Recall (KR), Exact Match (EM), and
Value~F1 (VF1).

\medskip
\noindent\textbf{Key Recall (KR):} Let

\[
K=\{k_i\}_{i=1}^{N}
\]

denote the collection of normalized ground-truth key occurrences, and let
$\hat{K}$ denote the corresponding collection of normalized predicted keys.
Accounting for duplicate occurrences, Key Recall is

\[
\mathrm{KR}
=
\frac{
\sum_{k}
\min\!\left(c_{K}(k),c_{\hat{K}}(k)\right)
}{N},
\]

where $c_K(k)$ and $c_{\hat K}(k)$ denote the number of occurrences of
normalized key $k$ in the ground truth and prediction, respectively.

\medskip
\noindent\textbf{Exact Match (EM):} A ground-truth pair receives exact-match credit only when an unused prediction
has both the same normalized key and an exactly matching normalized value.
Each prediction can be matched at most once. Document-level EM is

\[
\mathrm{EM}
=
\frac{N_{\mathrm{exact}}}{N},
\]

where $N_{\mathrm{exact}}$ is the number of ground-truth pairs for which such
an exact unused prediction exists.

\medskip
\noindent\textbf{Value F1 (VF1):} Value~F1 measures token-level overlap between a predicted value and its
reference value. Let $c_{\mathrm{pred}}(t)$ and $c_{\mathrm{gt}}(t)$ denote
the multiplicities of token $t$ in the predicted and ground-truth values. Token
overlap is

\[
O
=
\sum_t
\min\!\left(
c_{\mathrm{pred}}(t),
c_{\mathrm{gt}}(t)
\right).
\]

Token-level precision and recall are

\[
P
=
\frac{O}{|V_{\mathrm{pred}}|},
\qquad
R
=
\frac{O}{|V_{\mathrm{gt}}|},
\]

where $|V_{\mathrm{pred}}|$ and $|V_{\mathrm{gt}}|$ denote the numbers of
tokens in the predicted and reference values, including repeated tokens.
The corresponding score is

\[
F_1
=
\frac{2PR}{P+R},
\]

with $F_1=0$ when there is no token overlap.

For every ground-truth pair, an unused prediction with the same normalized key
is selected to maximize token-level F1. If no corresponding prediction exists,
that pair receives $F_{1,i}=0$. Document-level Value~F1 is therefore

\[
\mathrm{VF1}
=
\frac{1}{N}
\sum_{i=1}^{N}F_{1,i}.
\]

Importantly, the denominator is the total number of annotated ground-truth
pairs rather than only the subset of recovered keys. Consequently, missing
fields remain in the denominator and receive zero credit.

Thus, KR measures annotated-field recovery, EM complete KVP correctness, and
VF1 partial value recovery while retaining missing annotated pairs in the
denominator.

\subsection{Macro Averaging}

All reported dataset-level scores are macro-averaged over evaluable documents.
For metric $m$,

\[
\mathrm{Macro}(m)
=
\frac{1}{M}
\sum_{d=1}^{M}m_d,
\]

where $M$ denotes the number of evaluable documents and $m_d$ is the
document-level metric.

Macro averaging gives each evaluable document equal weight regardless of its
number of annotated KVPs; documents with no evaluable ground-truth KVPs are
excluded before aggregation.

\subsection{Evaluation Tracks}

Gold inputs use the deterministic annotation-derived representations in
Section~\ref{sec:text-input}; OCR inputs use PaddleOCR, EasyOCR, and Tesseract.
Comparisons hold the model, prompt structure, parsing, and evaluation fixed so
that the evaluated-document text source is the changing factor.

\subsection{Design Choices and Scope of the Metrics}

The protocol is intentionally reference-centered: fields absent from the
annotation schema are not automatically counted as false positives. This is
important for datasets such as SROIE, whose annotations cover only a subset of
semantically meaningful receipt content. Consequently, KR, EM, and VF1 should
not be interpreted as hallucination-frequency measures; unsupported or
additional fields are examined descriptively.

Duplicate semantic keys are evaluated as occurrences rather than unique names,
and predictions cannot be reused. Token-level VF1 uses token multiplicities
rather than set intersection, preserving repeated tokens while assigning zero
to unrecovered annotated pairs.

\subsection{Reproducibility and Evaluation Validation}

Raw generations, parsed predictions, and per-document scores are retained for
recomputation and inspection. Metric sanity checks cover perfect extraction,
correct keys with zero value overlap, missing keys, and mixtures of exact and
missing pairs, verifying in particular that missing ground-truth pairs remain
in the denominator.

For every document, the pipeline also records prompt length, input truncation,
generated-token count, and output-limit events, allowing context-budget and
generation-limit effects to be audited without post-hoc modification of
predictions or scores.

Implementation code for prompt construction, model inference, output parsing,
metric computation, and experiment execution is publicly available at
\url{https://github.com/zanvari/pixels-to-pairs}. The repository supports
reproduction and extension of the benchmark. Additional implementation
details are provided in Appendix A.

\section{Results}
\label{sec:results}

This section reports the quantitative results on FUNSD, SROIE, and CORD under annotation-derived Gold text and OCR-derived inputs. We report macro-averaged Key Recall (KR), Exact Match (EM), and Value~F1 (VF1) according to Section~\ref{sec:evaluation}. Unless otherwise specified, comparisons across model families use zero-shot prompting; one- to three-shot experiments are reported separately for Qwen2.5--7B and LLaMA~3--8B.

%


\begin{figure*}[t]
\centering
\setlength{\fboxsep}{1pt}
\setlength{\tabcolsep}{0pt}

\begin{minipage}[t]{0.485\linewidth}

\begin{minipage}[c][8mm][c]{\linewidth}
\centering
\fontsize{9}{11}\selectfont
\textbf{(a) Key-Label Corruption (FUNSD)}
\end{minipage}

\begin{minipage}[c][2.65cm][c]{\linewidth}
\centering
\fbox{\includegraphics[
    width=0.98\linewidth,
    height=2.50cm,
    keepaspectratio,
    trim={115 670 35 125},
    clip
]{funsd_87137840.png}}
\end{minipage}

\vspace{1.5mm}

{\fontsize{6.6}{7.7}\selectfont
\renewcommand{\arraystretch}{1.12}
\begin{tabularx}{\linewidth}{
    @{}>{\raggedright\arraybackslash}X
    @{\hspace{1.5mm}}
    >{\raggedright\arraybackslash}X@{}
}
\textbf{Gold excerpt} & \textbf{PaddleOCR excerpt} \\[-0.2mm]
\hline
\texttt{TA \#: T07281A}\par
\texttt{SPONSOR'S NAME: Lorillard}\par
\texttt{TEST ARTICLE IDENTIFICATION: B220}\par
\texttt{LOT \#: 012590}\par
\texttt{CONDITION OF SHIPMENT:}\par
\texttt{GOOD BROKEN LEAKED}
&
\textcolor{red!70!black}{\texttt{[TA \# label omitted]}}
\texttt{ T07281A}\par
\texttt{SPONSOR'S NAME: }
\textcolor{red!70!black}{\texttt{LORIllard}}\par
\textcolor{red!70!black}{\texttt{TEST ARTTCKE}}
\texttt{ IDENTIFICATION: : B220}\par
\textcolor{red!70!black}{\texttt{LoT}}
\texttt{ \#: 012590}\par
\textcolor{red!70!black}{\texttt{CONDTTION OF SHIMENT:}}\par
\textcolor{red!70!black}{\texttt{OODEROKEN IZAFED}}
\end{tabularx}\par}

\vspace{1mm}

{\raggedright\fontsize{7.1}{8.5}\selectfont
\textbf{Extraction impact:} Qwen omits \texttt{TA \#}, reproduces the
corrupted sponsor and condition values, and preserves \texttt{012590}
under the OCR-corrupted \texttt{LoT \#} label.\par}

\end{minipage}
\hfill
\begin{minipage}[t]{0.485\linewidth}

\begin{minipage}[c][8mm][c]{\linewidth}
\centering
\fontsize{9}{11}\selectfont
\textbf{(b) Character-Level Value Corruption (SROIE)}
\end{minipage}

\begin{minipage}[c][2.65cm][c]{\linewidth}
\centering
\fbox{\includegraphics[
    width=0.98\linewidth,
    height=2.50cm,
    keepaspectratio,
    trim={35 1435 35 120},
    clip
]{sroie_X51006328937.jpg}}
\end{minipage}

\vspace{1.5mm}

{\fontsize{6.6}{7.7}\selectfont
\renewcommand{\arraystretch}{1.12}
\begin{tabularx}{\linewidth}{
    @{}>{\raggedright\arraybackslash}X
    @{\hspace{1.5mm}}
    >{\raggedright\arraybackslash}X@{}
}
\textbf{Gold excerpt} & \textbf{PaddleOCR excerpt} \\[-0.2mm]
\hline
\texttt{OLIVE9 PHARMACY SDN BHD}\par
\texttt{1030214K}\par
\texttt{GST/TAX NO: 000988880896}\par
\texttt{69 JALAN NOVA U5/N, SEKSYEN U5 SUBANG BE}\par
\texttt{TEL: 03-77347762 FAX: 03-78327762}\par
\texttt{GST NO: 000988880896}
&
\textcolor{red!70!black}{\texttt{YLIVE9}}
\texttt{ PHARMACY SDN BHD}\par
\texttt{1030214K}\par
\texttt{GST/TAX NO: 000988880896}\par
\texttt{69 JALAN NOVA }
\textcolor{red!70!black}{\texttt{US/N. EBESYEN US SUBANG BB}}\par
\textcolor{red!70!black}{\texttt{TBL:}}
\texttt{ 03-77347762 FAX: 03-78327762}\par
\texttt{GST NO: 000988880896}
\end{tabularx}\par}

\vspace{1mm}

{\raggedright\fontsize{7.1}{8.5}\selectfont
\textbf{Extraction impact:} Qwen preserves the relevant field structure
but propagates the OCR-corrupted company and address values.\par}

\end{minipage}

\par\vspace{4mm}

\begin{minipage}[t]{0.485\linewidth}

\begin{minipage}[c][8mm][c]{\linewidth}
\centering
\fontsize{9}{11}\selectfont
\textbf{(c) Cascading Value Misalignment (CORD)}
\end{minipage}

\begin{minipage}[c][2.95cm][c]{\linewidth}
\centering
\fbox{\includegraphics[
    width=0.98\linewidth,
    height=2.80cm,
    keepaspectratio,
    trim={100 311 85 610},
    clip
]{cord_receipt_00045.png}}
\end{minipage}

\vspace{1.5mm}

{\fontsize{6.6}{7.7}\selectfont
\renewcommand{\arraystretch}{1.12}
\begin{tabularx}{\linewidth}{
    @{}>{\raggedright\arraybackslash}X
    @{\hspace{1.5mm}}
    >{\raggedright\arraybackslash}X@{}
}
\textbf{Gold excerpt} & \textbf{PaddleOCR excerpt} \\[-0.2mm]
\hline
\texttt{1 $\mid$ Lemon Tea (L.) $\mid$ 25.000}\par
\texttt{1 $\mid$ Mike \& Ike Original $\mid$ 28.000}\par
\texttt{1 $\mid$ Mike \& Berry $\mid$ 28.000}\par
\texttt{1 $\mid$ Roti Keju Manis. $\mid$ 7.000}\par
\texttt{TOTAL $\mid$ 88,000}
&
\textcolor{red!70!black}{\texttt{25.000 $\mid$ 1}}
\texttt{ $\mid$ Lemon Tea (L.) $\mid$ 28.000}\par
\texttt{Mike \& Ike Original $\mid$ 28.000}\par
\texttt{Mike \& Berry $\mid$ 7.000}\par
\texttt{Roti Keju Manis. $\mid$ 88.000 $\mid$ TOTAL}
\end{tabularx}\par}

\vspace{1mm}

{\raggedright\fontsize{7.1}{8.5}\selectfont
\textbf{Extraction impact:} The displaced value sequence propagates
across the item list, corrupting three item values and the total
formatting.\par}

\end{minipage}
\hfill
\begin{minipage}[t]{0.485\linewidth}

\begin{minipage}[c][8mm][c]{\linewidth}
\centering
\fontsize{9}{11}\selectfont
\textbf{(d) OCR-Induced Error Amplification (SROIE)}
\end{minipage}

{\begin{minipage}[t][2.85cm][c]{\linewidth}
\centering
\fbox{\includegraphics[
    width=0.98\linewidth,
    height=2.80cm,
    keepaspectratio,
    trim={25 1090 25 260},
    clip
]{sroie_X51006619784.jpg}}\par
\end{minipage}}

\vspace{1.5mm}

{\fontsize{6.6}{7.7}\selectfont
\renewcommand{\arraystretch}{1.12}
\begin{tabularx}{\linewidth}{
    @{}>{\raggedright\arraybackslash}X
    @{\hspace{1.5mm}}
    >{\raggedright\arraybackslash}X@{}
}
\textbf{Gold excerpt} & \textbf{PaddleOCR excerpt} \\[-0.2mm]
\hline
\texttt{SECURITY \& OA TRADING}\par
\texttt{GST NO: 000345100288}\par
\texttt{DATE: 7/6/2016}\par
\texttt{TIME: 2:47:04 PM}
&
\texttt{SECURITY \&OA TRADING:}\par
\texttt{GST NO: 000345100288}\par
\texttt{Date: }
\textcolor{red!70!black}{\texttt{7/0z018}}\par
\texttt{Time: 2:47:04 PM}
\end{tabularx}\par}

\vspace{1mm}

{\raggedright\fontsize{7.1}{8.5}\selectfont
\textbf{Extraction impact:} Gold-text inference returns
\texttt{date} $\rightarrow$ \texttt{7/6/2016}. From the malformed
PaddleOCR string \texttt{7/0z018}, Qwen instead returns
\texttt{date} $\rightarrow$ \texttt{7/02/2018}, converting the OCR
error into a plausible-looking but incorrect date.\par}

\end{minipage}

\par\vspace{2mm}

\caption{Four qualitative error-propagation pathways for zero-shot
Qwen2.5--7B under PaddleOCR-derived text:
(a) \textbf{key-label loss and corruption} in FUNSD document 87137840;
(b) \textbf{character-level value corruption} in SROIE document
X51006328937;
(c) \textbf{reading-order displacement with cascading value
misalignment} in CORD document receipt\_00045; and
(d) \textbf{OCR-induced error amplification} in SROIE document
X51006619784. Each panel presents a focused document crop, the
corresponding Gold and PaddleOCR excerpts, and the downstream extraction
impact. Red marks the relevant OCR-induced changes.}
\label{fig:qualitative_examples}
\end{figure*}

\subsection{Overall Findings}
The results reveal three broad patterns. First, extraction difficulty varies substantially across document domains. Zero-shot Gold-text KR ranges from 0.2763--0.6332 across models on FUNSD, whereas several models approach complete key recovery on SROIE. Second, OCR-derived text generally produces substantial reductions in EM and VF1, although the magnitude of degradation depends on the dataset, OCR engine, and model. Third, in-context demonstrations can produce large gains for some model--dataset combinations, but these gains are neither monotonic in the number of demonstrations nor consistent across models.

These observations indicate that downstream KVP extraction depends jointly on document structure, upstream text quality, model family, and demonstration configuration rather than on any single factor in isolation.

\subsection{OCR Quality Across Datasets}
Before examining downstream extraction, Table~\ref{tab:ocr-error-rates} summarizes OCR quality using Character Error Rate (CER) and Word Error Rate (WER). Lower values indicate greater similarity to the corresponding reference text.

\begin{table}[t]
\centering
\setlength{\tabcolsep}{4pt}
\caption{Character Error Rate (CER) and Word Error Rate (WER) for the evaluated OCR engines across the three benchmark datasets. Lower values indicate better OCR quality.}
\label{tab:ocr-error-rates}
\begin{tabular}{@{}llcc@{}}
\toprule
\textbf{Dataset} & \textbf{OCR Engine} & \textbf{Mean CER} & \textbf{Mean WER} \\
\midrule
\multirow{3}{*}{FUNSD}
 & PaddleOCR & \textbf{0.4199} & \textbf{0.5675} \\
 & EasyOCR   & 0.4974 & 0.7318 \\
 & Tesseract & 0.5485 & 0.7197 \\
\midrule
\multirow{3}{*}{SROIE}
 & PaddleOCR & \textbf{0.3099} & \textbf{0.4577} \\
 & EasyOCR   & 0.4145 & 0.7148 \\
 & Tesseract & 0.4334 & 0.6615 \\
\midrule
\multirow{3}{*}{CORD}
 & PaddleOCR & \textbf{0.4844} & \textbf{0.6199} \\
 & EasyOCR   & 0.5044 & 0.8519 \\
 & Tesseract & 0.8031 & 0.9825 \\
\bottomrule
\end{tabular}
\end{table}

PaddleOCR has the lowest CER and WER on all three datasets. SROIE exhibits the lowest OCR error rates overall, whereas CORD is particularly challenging under Tesseract. CER and WER, however, do not uniquely determine downstream extraction performance because recognition errors differ in their location and semantic importance.

These are sequence-level error rates between each engine’s native plain-text serialization and the corresponding annotation-derived reference, rather than isolated glyph-recognition error rates. Particularly on FUNSD, their magnitude reflects not only character substitutions and omissions but also differences in line grouping and reading order relative to the spatially reconstructed Gold text.
\subsection{FUNSD Results}
\label{sec:results-funsd}
FUNSD is particularly challenging in the text-only setting because many question--answer relationships depend on two-dimensional layout. The corrected Gold-text representation described in Section~\ref{sec:text-input} spatially reconstructs annotation text before flattening it into a sequence.

\begin{table}[t]
\centering
\footnotesize
\setlength{\tabcolsep}{3pt}
\renewcommand{\arraystretch}{0.96}
\caption{Zero-shot KVP extraction performance on \textbf{FUNSD} across
Gold and OCR-derived text. Bold values indicate the best score
within each text condition.}
\label{tab:funsd-zero-shot}
\begin{tabular}{@{}llccc@{}}
\toprule
\textbf{Text} & \textbf{Model} &
\textbf{KR} & \textbf{EM} & \textbf{VF1} \\
\midrule

\multirow{5}{*}{Gold}
& Gemma 2B       & 0.2763 & 0.2252 & 0.2404 \\
& Gemma 7B       & 0.3990 & 0.3080 & 0.3433 \\
& Mistral--7B    & 0.5850 & 0.4627 & 0.5102 \\
& Qwen2.5--7B    & \textbf{0.6332} & \textbf{0.5040} & \textbf{0.5519} \\
& LLaMA~3--8B    & 0.4384 & 0.3032 & 0.3594 \\
\midrule

\multirow{5}{*}{PaddleOCR}
& Gemma 2B       & 0.3067 & 0.1352 & 0.1887 \\
& Gemma 7B       & 0.5040 & 0.2211 & 0.3087 \\
& Mistral--7B    & \textbf{0.6403} & \textbf{0.2924} & \textbf{0.3690} \\
& Qwen2.5--7B    & 0.5703 & 0.2477 & 0.3532 \\
& LLaMA~3--8B    & 0.4653 & 0.2072 & 0.2848 \\
\midrule

\multirow{5}{*}{EasyOCR}
& Gemma 2B       & 0.1613 & 0.0296 & 0.0586 \\
& Gemma 7B       & 0.3211 & 0.0574 & 0.1186 \\
& Mistral--7B    & \textbf{0.3897} & \textbf{0.0628} & \textbf{0.1433} \\
& Qwen2.5--7B    & 0.3841 & 0.0582 & 0.1405 \\
& LLaMA~3--8B    & 0.2705 & 0.0394 & 0.0952 \\
\midrule

\multirow{5}{*}{Tesseract}
& Gemma 2B       & 0.1760 & 0.0598 & 0.0941 \\
& Gemma 7B       & 0.2008 & 0.0409 & 0.0712 \\
& Mistral--7B    & \textbf{0.3704} & 0.1057 & 0.1775 \\
& Qwen2.5--7B    & 0.3546 & \textbf{0.1080} & \textbf{0.1784} \\
& LLaMA~3--8B    & 0.3002 & 0.0684 & 0.1230 \\
\bottomrule
\end{tabular}
\end{table}

Table~\ref{tab:funsd-zero-shot} reports zero-shot performance
across Gold and OCR-derived text for the 47 evaluable FUNSD documents.

Qwen2.5--7B obtains the highest zero-shot KR, EM, and VF1 on the corrected Gold-text representation. Mistral--7B is the next strongest zero-shot model on all three metrics. The remaining gap between KR and EM across models indicates that detecting an annotated field does not guarantee recovery of its exact associated value.

\subsubsection{Few-Shot Prompting on Gold Text}
Figure~\ref{fig:few_shot_trajectories}(a) shows the zero-
to three-shot trajectories under FUNSD Gold text; the
corresponding numerical results are consolidated in
Table~\ref{tab:combined_fewshot}.

The two models respond differently to demonstrations. LLaMA~3--8B improves substantially after adding demonstrations and reaches its highest scores at two shots. Qwen2.5--7B, in contrast, does not improve monotonically: zero-shot prompting gives its highest KR, while three-shot prompting gives slightly higher EM and VF1. These results show that increasing the number of demonstrations does not necessarily produce monotonic improvements.

\subsubsection{OCR-Derived Inputs}
The OCR-derived results in Table~\ref{tab:funsd-zero-shot}
show substantial differences among PaddleOCR, EasyOCR, and Tesseract.

PaddleOCR produces substantially stronger downstream results than EasyOCR and Tesseract for most configurations. Importantly, the relative ordering of models also changes across text conditions. Mistral--7B obtains the highest zero-shot KR, EM, and VF1 under PaddleOCR despite Qwen2.5--7B leading on Gold text. Under Tesseract, Mistral retains the highest KR, whereas Qwen2.5--7B has the highest EM and VF1. Thus, model rankings under clean text do not necessarily transfer directly to OCR-derived inputs.

\subsubsection{Few-Shot Prompting under PaddleOCR}

\begin{figure*}[t]
\centering

\newcommand{\plotfewshotseries}[6]{%
    \addplot[qwenKR]  coordinates {#1};
    \addplot[qwenEM]  coordinates {#2};
    \addplot[qwenVF]  coordinates {#3};
    \addplot[llamaKR] coordinates {#4};
    \addplot[llamaEM] coordinates {#5};
    \addplot[llamaVF] coordinates {#6};
}

\begin{tikzpicture}
\begin{groupplot}[
    group style={
    group size=2 by 3,
    horizontal sep=1.0cm,
    vertical sep=1.15cm,
    xlabels at=edge bottom,
    xticklabels at=edge bottom,
    ylabels at=edge left,
    yticklabels at=edge left
},
    width=0.455\textwidth,
    height=3.6cm,
    xmin=-0.05,
    xmax=3.05,
    xtick={0,1,2,3},
    xmajorgrids=true,
    ymajorgrids=true,
    grid style={black!15},
    axis lines*=left,
    axis line style={black!70},
    tick label style={font=\scriptsize},
    label style={font=\small},
    title style={
    font=\small\bfseries,
    at={(axis description cs:0.5,1.03)},
    anchor=south
},
    scaled ticks=false,
]

\nextgroupplot[
    title={(a) FUNSD -- Gold text},
    ylabel={Score (\%)},
    ymin=15,
    ymax=70,
    ytick={20,30,40,50,60,70},
    legend to name=fewshotlegend,
    legend columns=3,
    legend cell align={left},
    legend style={
        font=\scriptsize,
        draw=none,
        /tikz/every even column/.append style={column sep=0.6em}
    }
]
\plotfewshotseries
    {(0,63.32) (1,60.43) (2,60.23) (3,62.23)}
    {(0,50.40) (1,49.91) (2,49.63) (3,51.10)}
    {(0,55.19) (1,54.26) (2,54.06) (3,56.03)}
    {(0,43.84) (1,63.19) (2,66.57) (3,64.51)}
    {(0,30.32) (1,52.54) (2,54.67) (3,54.47)}
    {(0,35.94) (1,55.67) (2,58.72) (3,58.10)}

\addlegendimage{qwenKR}
\addlegendentry{Qwen--KR}
\addlegendimage{qwenEM}
\addlegendentry{Qwen--EM}
\addlegendimage{qwenVF}
\addlegendentry{Qwen--VF1}
\addlegendimage{llamaKR}
\addlegendentry{LLaMA--KR}
\addlegendimage{llamaEM}
\addlegendentry{LLaMA--EM}
\addlegendimage{llamaVF}
\addlegendentry{LLaMA--VF1}

\nextgroupplot[
    title={(b) FUNSD -- PaddleOCR},
    ymin=15,
    ymax=70,
    ytick={20,30,40,50,60,70}
]
\plotfewshotseries
    {(0,57.03) (1,59.41) (2,57.02) (3,60.35)}
    {(0,24.77) (1,26.72) (2,24.92) (3,28.52)}
    {(0,35.32) (1,36.58) (2,33.58) (3,38.24)}
    {(0,46.53) (1,62.71) (2,63.18) (3,63.47)}
    {(0,20.72) (1,26.58) (2,28.35) (3,27.54)}
    {(0,28.48) (1,37.07) (2,39.21) (3,38.43)}

\nextgroupplot[
    title={(c) SROIE -- Gold text},
    ylabel={Score (\%)},
    ymin=30,
    ymax=102,
    ytick={40,60,80,100}
]
\plotfewshotseries
    {(0,99.50) (1,99.06) (2,99.93) (3,99.71)}
    {(0,68.66) (1,73.70) (2,76.22) (3,76.80)}
    {(0,89.30) (1,88.90) (2,90.50) (3,90.44)}
    {(0,65.42) (1,99.78) (2,99.78) (3,99.78)}
    {(0,38.69) (1,71.33) (2,79.25) (3,78.60)}
    {(0,55.10) (1,86.79) (2,89.39) (3,89.40)}

\nextgroupplot[
    title={(d) SROIE -- PaddleOCR},
    ymin=30,
    ymax=102,
    ytick={40,60,80,100}
]
\plotfewshotseries
    {(0,99.35) (1,99.14) (2,99.64) (3,99.35)}
    {(0,57.42) (1,63.90) (2,66.07) (3,67.15)}
    {(0,83.67) (1,83.71) (2,85.39) (3,85.61)}
    {(0,68.37) (1,98.85) (2,99.71) (3,99.78)}
    {(0,34.44) (1,59.87) (2,68.59) (3,68.37)}
    {(0,55.00) (1,82.19) (2,85.08) (3,85.14)}

\nextgroupplot[
    title={(e) CORD -- Gold text},
    xlabel={Number of shots},
    ylabel={Score (\%)},
    ymin=35,
    ymax=95,
    ytick={40,50,60,70,80,90}
]
\plotfewshotseries
    {(0,90.45) (1,92.05) (2,91.19) (3,89.96)}
    {(0,87.36) (1,89.51) (2,89.31) (3,87.57)}
    {(0,87.48) (1,89.59) (2,89.31) (3,87.70)}
    {(0,85.81) (1,89.43) (2,90.85) (3,90.25)}
    {(0,73.83) (1,81.49) (2,83.08) (3,81.67)}
    {(0,73.83) (1,81.60) (2,83.33) (3,81.91)}

\nextgroupplot[
    title={(f) CORD -- PaddleOCR},
    xlabel={Number of shots},
    ymin=35,
    ymax=95,
    ytick={40,50,60,70,80,90}
]
\plotfewshotseries
    {(0,69.27) (1,72.53) (2,70.41) (3,69.81)}
    {(0,48.55) (1,50.90) (2,48.06) (3,48.60)}
    {(0,49.06) (1,51.82) (2,48.82) (3,49.33)}
    {(0,68.79) (1,70.08) (2,70.33) (3,72.05)}
    {(0,41.09) (1,45.26) (2,46.06) (3,47.61)}
    {(0,42.07) (1,45.77) (2,46.57) (3,47.68)}

\end{groupplot}
\end{tikzpicture}

\vspace{0.25em}
\pgfplotslegendfromname{fewshotlegend}

\caption{Few-shot extraction trajectories for Qwen2.5--7B
and LLaMA~3--8B across FUNSD, SROIE, and CORD under
annotation-derived Gold text and PaddleOCR-derived text.
Colors and marker shapes distinguish Key Recall (KR), Exact
Match (EM), and Value F1 (VF1); solid filled markers denote
Qwen2.5--7B, while dashed open markers denote LLaMA~3--8B.
The trajectories demonstrate that the effect of additional
demonstrations depends on the model, dataset, metric, and text
condition, with no uniformly optimal demonstration count.
Axis ranges are shared within each dataset row.}
\label{fig:few_shot_trajectories}
\end{figure*}
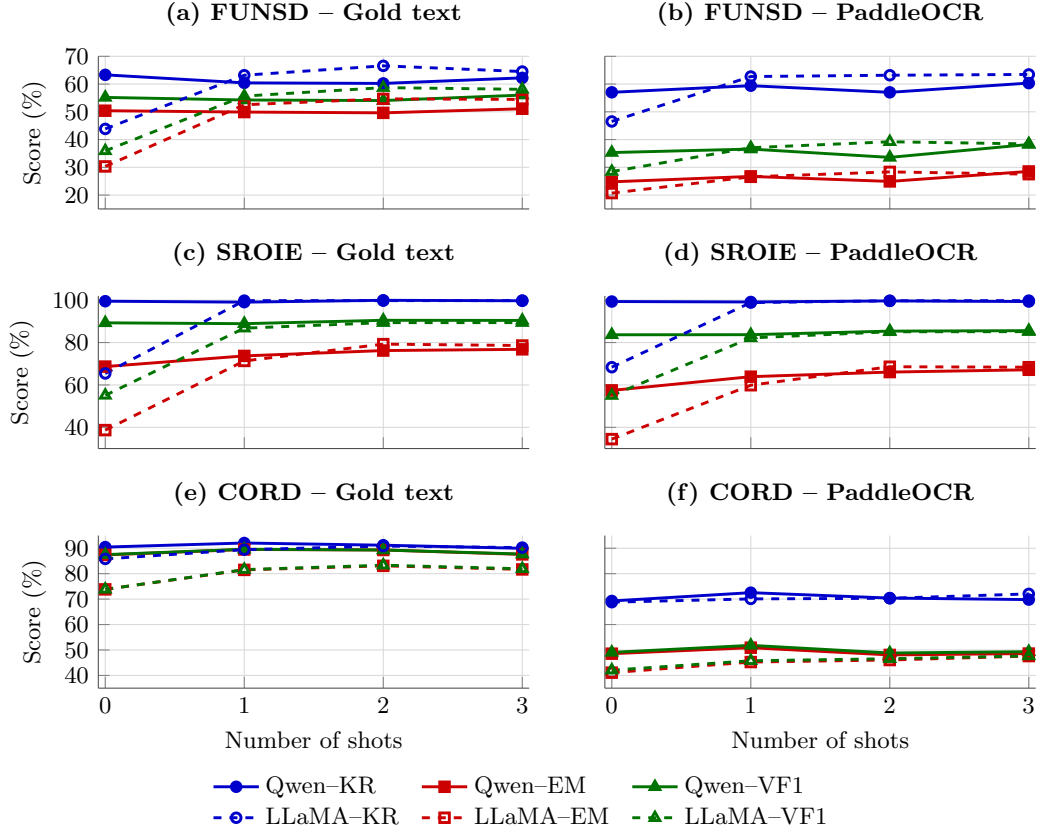

\begin{table*}[t]
\centering
\caption{Few-shot KVP extraction performance across datasets under
annotation-derived Gold text and PaddleOCR-derived text. Within each
dataset, the same clean training demonstrations were used under both
text conditions. Bold values indicate the best score for each
model, text condition, and metric across the evaluated shot counts.}
\label{tab:combined_fewshot}

\scriptsize
\setlength{\tabcolsep}{3.2pt}
\renewcommand{\arraystretch}{0.92}

\begin{tabular*}{\textwidth}{
    @{\extracolsep{\fill}}
    llc
    ccc
    ccc
    @{}
}
\toprule
& & &
\multicolumn{3}{c}{Gold Text} &
\multicolumn{3}{c}{PaddleOCR} \\
\cmidrule(lr){4-6}
\cmidrule(lr){7-9}
Dataset & Model & Shots &
KR & EM & VF1 &
KR & EM & VF1 \\
\midrule

\multirow{8}{*}{FUNSD}
& \multirow{4}{*}{Qwen2.5--7B}
& 0 & \textbf{0.6332} & 0.5040 & 0.5519
    & 0.5703 & 0.2477 & 0.3532 \\
&
& 1 & 0.6043 & 0.4991 & 0.5426
    & 0.5941 & 0.2672 & 0.3658 \\
&
& 2 & 0.6023 & 0.4963 & 0.5406
    & 0.5702 & 0.2492 & 0.3358 \\
&
& 3 & 0.6223 & \textbf{0.5110} & \textbf{0.5603}
    & \textbf{0.6035} & \textbf{0.2852} & \textbf{0.3824} \\
\cmidrule(lr){2-9}
&
\multirow{4}{*}{LLaMA~3--8B}
& 0 & 0.4384 & 0.3032 & 0.3594
    & 0.4653 & 0.2072 & 0.2848 \\
&
& 1 & 0.6319 & 0.5254 & 0.5567
    & 0.6271 & 0.2658 & 0.3707 \\
&
& 2 & \textbf{0.6657} & \textbf{0.5467} & \textbf{0.5872}
    & 0.6318 & \textbf{0.2835} & \textbf{0.3921} \\
&
& 3 & 0.6451 & 0.5447 & 0.5810
    & \textbf{0.6347} & 0.2754 & 0.3843 \\
\midrule

\multirow{8}{*}{SROIE}
& \multirow{4}{*}{Qwen2.5--7B}
& 0 & 0.9950 & 0.6866 & 0.8930
    & 0.9935 & 0.5742 & 0.8367 \\
&
& 1 & 0.9906 & 0.7370 & 0.8890
    & 0.9914 & 0.6390 & 0.8371 \\
&
& 2 & \textbf{0.9993} & 0.7622 & \textbf{0.9050}
    & \textbf{0.9964} & 0.6607 & 0.8539 \\
&
& 3 & 0.9971 & \textbf{0.7680} & 0.9044
    & 0.9935 & \textbf{0.6715} & \textbf{0.8561} \\
\cmidrule(lr){2-9}
&
\multirow{4}{*}{LLaMA~3--8B}
& 0 & 0.6542 & 0.3869 & 0.5510
    & 0.6837 & 0.3444 & 0.5500 \\
&
& 1 & \textbf{0.9978} & 0.7133 & 0.8679
    & 0.9885 & 0.5987 & 0.8219 \\
&
& 2 & \textbf{0.9978} & \textbf{0.7925} & 0.8939
    & 0.9971 & \textbf{0.6859} & 0.8508 \\
&
& 3 & \textbf{0.9978} & 0.7860 & \textbf{0.8940}
    & \textbf{0.9978} & 0.6837 & \textbf{0.8514} \\
\midrule

\multirow{8}{*}{CORD}
& \multirow{4}{*}{Qwen2.5--7B}
& 0 & 0.9045 & 0.8736 & 0.8748
    & 0.6927 & 0.4855 & 0.4906 \\
&
& 1 & \textbf{0.9205} & \textbf{0.8951} & \textbf{0.8959}
    & \textbf{0.7253} & \textbf{0.5090} & \textbf{0.5182} \\
&
& 2 & 0.9119 & 0.8931 & 0.8931
    & 0.7041 & 0.4806 & 0.4882 \\
&
& 3 & 0.8996 & 0.8757 & 0.8770
    & 0.6981 & 0.4860 & 0.4933 \\
\cmidrule(lr){2-9}
&
\multirow{4}{*}{LLaMA~3--8B}
& 0 & 0.8581 & 0.7383 & 0.7383
    & 0.6879 & 0.4109 & 0.4207 \\
&
& 1 & 0.8943 & 0.8149 & 0.8160
    & 0.7008 & 0.4526 & 0.4577 \\
&
& 2 & \textbf{0.9085} & \textbf{0.8308} & \textbf{0.8333}
    & 0.7033 & 0.4606 & 0.4657 \\
&
& 3 & 0.9025 & 0.8167 & 0.8191
    & \textbf{0.7205} & \textbf{0.4761} & \textbf{0.4768} \\
\bottomrule
\end{tabular*}
\end{table*}

Figure~\ref{fig:few_shot_trajectories}(b) shows the FUNSD
PaddleOCR trajectories, and Table~\ref{tab:combined_fewshot}
reports the corresponding scores using the same clean training
demonstrations as in the Gold-text experiment.


The trajectories show that additional demonstrations do not
consistently improve extraction and can instead reduce it.
Qwen2.5--7B obtains its highest PaddleOCR scores at three
shots, but its two-shot VF1 falls below both one-shot and
zero-shot performance. LLaMA~3--8B obtains its highest KR at
three shots but its highest EM and VF1 at two shots. These
results reinforce that demonstration count alone is not a
reliable predictor of extraction quality.

\subsection{Sensitivity to Demonstration Composition}
\label{sec:results-demo-sensitivity}

\begin{figure}[!t]
\centering
\begin{tikzpicture}
\begin{axis}[
    boxplot/draw direction=y,
    xtick={1,2,3,4},
    xticklabels={Low, Representative, High, Diverse},
    x tick label style={font=\footnotesize, rotate=15, anchor=north east},
    ylabel={Value F1 Score (\%)},
    width=\linewidth, 
    height=5.5cm,
    ymin=45, ymax=65,
    ymajorgrids=true,
    axis lines*=left,
    label style={font=\footnotesize}
]
\addplot+[
    boxplot prepared={
        lower whisker=48.45, lower quartile=50.00,
        median=51.46, upper quartile=52.50, upper whisker=53.99
    },
    draw=black, thick, fill=blue!20
] coordinates {};
\addplot+[
    boxplot prepared={
        lower whisker=53.73, lower quartile=55.50,
        median=57.85, upper quartile=59.50, upper whisker=61.76
    },
    draw=black, thick, fill=red!20
] coordinates {};
\addplot+[
    boxplot prepared={
        lower whisker=52.97, lower quartile=54.00,
        median=55.06, upper quartile=56.00, upper whisker=56.74
    },
    draw=black, thick, fill=green!20
] coordinates {};
\addplot+[
    boxplot prepared={
        lower whisker=53.18, lower quartile=53.80,
        median=54.36, upper quartile=55.00, upper whisker=55.95
    },
    draw=black, thick, fill=orange!20
] coordinates {};
\end{axis}
\end{tikzpicture}
\caption{Confirmatory Sensitivity Analysis: Variability in 3-shot Value F1 performance across demonstration complexity strategies (Qwen2.5-7B on FUNSD Gold). The observed ranges—for example, $53.73\%$ to $61.76\%$ within the Representative strategy—demonstrate substantial sensitivity to demonstration composition even when demonstration count and selection strategy are fixed.}
\label{fig:prompt_sensitivity}
\end{figure}
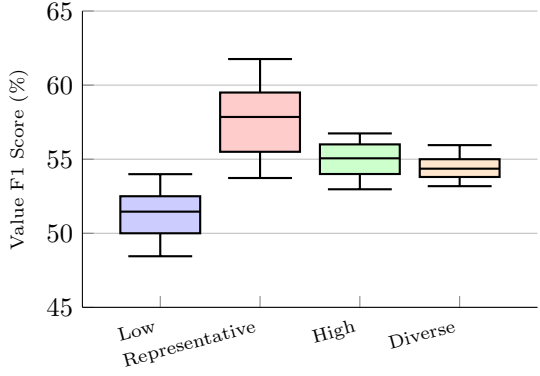

Following the protocol in Section~\ref{sec:demo-sensitivity-method}, we evaluated 20 frozen three-shot demonstration sets---five each under the Low, Representative, High, and Diverse strategies---on FUNSD Gold text with Qwen2.5--7B and LLaMA~3--8B, yielding 40 sensitivity configurations. The 60 training documents were not reused across sets, and all sets were frozen before inference without model-output or test-performance feedback.

A context audit found no truncation in the main FUNSD runs or in 38 of the 40 sensitivity configurations. Qwen High-4 truncated the target document in all 47 cases and was excluded from aggregation; LLaMA High-4 truncated 2/47 documents and was retained. Table~\ref{tab:funsd-demo-sensitivity} summarizes the valid configurations.

\begin{table*}[t]
\centering
\caption{Sensitivity of three-shot FUNSD Gold-text performance to demonstration composition. Each cell reports mean $\pm$ sample standard deviation, with the observed range shown underneath.}
\label{tab:funsd-demo-sensitivity}
\begingroup
\footnotesize
\setlength{\tabcolsep}{2.5pt}
\renewcommand{\arraystretch}{1.18}
\begin{tabularx}{\textwidth}{
@{}
L{1.65cm}
L{2.05cm}
c
>{\centering\arraybackslash}X
>{\centering\arraybackslash}X
>{\centering\arraybackslash}X
@{}
}
\toprule
\textbf{Model} &
\textbf{Strategy} &
\textbf{$n$} &
\textbf{KR} &
\textbf{EM} &
\textbf{VF1} \\
\midrule
\multirow{4}{1.65cm}{Qwen2.5-7B}
& Low & 5 &
\shortstack{$0.5790 \pm 0.0219$\\[-1pt]
{\scriptsize $[0.5514,\,0.6119]$}} &
\shortstack{$0.4659 \pm 0.0242$\\[-1pt]
{\scriptsize $[0.4272,\,0.4906]$}} &
\shortstack{$0.5146 \pm 0.0197$\\[-1pt]
{\scriptsize $[0.4845,\,0.5399]$}} \\
& Representative & 5 &
\shortstack{$0.6530 \pm 0.0329$\\[-1pt]
{\scriptsize $[0.6169,\,0.6932]$}} &
\shortstack{$0.5437 \pm 0.0455$\\[-1pt]
{\scriptsize $[0.4933,\,0.5961]$}} &
\shortstack{$0.5785 \pm 0.0374$\\[-1pt]
{\scriptsize $[0.5373,\,0.6176]$}} \\
& High & 4 &
\shortstack{$0.6301 \pm 0.0075$\\[-1pt]
{\scriptsize $[0.6229,\,0.6395]$}} &
\shortstack{$0.5072 \pm 0.0242$\\[-1pt]
{\scriptsize $[0.4754,\,0.5276]$}} &
\shortstack{$0.5506 \pm 0.0161$\\[-1pt]
{\scriptsize $[0.5297,\,0.5674]$}} \\
& Diverse & 5 &
\shortstack{$0.6190 \pm 0.0204$\\[-1pt]
{\scriptsize $[0.6001,\,0.6458]$}} &
\shortstack{$0.5031 \pm 0.0202$\\[-1pt]
{\scriptsize $[0.4847,\,0.5341]$}} &
\shortstack{$0.5436 \pm 0.0140$\\[-1pt]
{\scriptsize $[0.5318,\,0.5595]$}} \\
\midrule
\multirow{4}{1.65cm}{LLaMA-3-8B}
& Low & 5 &
\shortstack{$0.6130 \pm 0.0302$\\[-1pt]
{\scriptsize $[0.5784,\,0.6558]$}} &
\shortstack{$0.5168 \pm 0.0227$\\[-1pt]
{\scriptsize $[0.4928,\,0.5440]$}} &
\shortstack{$0.5531 \pm 0.0203$\\[-1pt]
{\scriptsize $[0.5313,\,0.5824]$}} \\
& Representative & 5 &
\shortstack{$0.6578 \pm 0.0315$\\[-1pt]
{\scriptsize $[0.6254,\,0.7095]$}} &
\shortstack{$0.5459 \pm 0.0349$\\[-1pt]
{\scriptsize $[0.5069,\,0.6000]$}} &
\shortstack{$0.5829 \pm 0.0297$\\[-1pt]
{\scriptsize $[0.5530,\,0.6321]$}} \\
& High & 5 &
\shortstack{$0.6410 \pm 0.0397$\\[-1pt]
{\scriptsize $[0.5799,\,0.6738]$}} &
\shortstack{$0.5343 \pm 0.0344$\\[-1pt]
{\scriptsize $[0.4861,\,0.5723]$}} &
\shortstack{$0.5700 \pm 0.0330$\\[-1pt]
{\scriptsize $[0.5224,\,0.6044]$}} \\
& Diverse & 5 &
\shortstack{$0.6542 \pm 0.0184$\\[-1pt]
{\scriptsize $[0.6214,\,0.6639]$}} &
\shortstack{$0.5436 \pm 0.0233$\\[-1pt]
{\scriptsize $[0.5109,\,0.5669]$}} &
\shortstack{$0.5823 \pm 0.0212$\\[-1pt]
{\scriptsize $[0.5466,\,0.6004]$}} \\
\bottomrule
\end{tabularx}
\endgroup
\end{table*}

Low-complexity sets yield the lowest mean VF1 for both models: 0.5146 for Qwen2.5--7B and 0.5531 for LLaMA~3--8B. Representative sets produce the highest means (0.5785 and 0.5829, respectively), although LLaMA obtains a nearly identical mean under Diverse demonstrations (0.5823), confirming that increasing demonstration complexity is not monotonically beneficial.

Substantial within-strategy variation also remains. Across Representative sets, VF1 ranges from 0.5373 to 0.6176 for Qwen2.5--7B and from 0.5530 to 0.6321 for LLaMA~3--8B. Figure~\ref{fig:prompt_sensitivity} visualizes the Qwen variation, demonstrating that conclusions based on a single demonstration set can depend materially on the selected examples.


\subsection{SROIE Results}
\label{sec:results-sroie}

SROIE evaluates receipt extraction using a compact annotation schema. The dataset is substantially easier for key identification than FUNSD under the text-only setting, with several models approaching complete key recovery.

\begin{table}[t]
\centering
\footnotesize
\setlength{\tabcolsep}{3pt}
\renewcommand{\arraystretch}{0.96}
\caption{Zero-shot KVP extraction performance on \textbf{SROIE} across
Gold and OCR-derived text. Bold values indicate the best score
within each text condition.}
\label{tab:sroie-zero-shot}
\begin{tabular}{@{}llccc@{}}
\toprule
\textbf{Text} & \textbf{Model} &
\textbf{KR} & \textbf{EM} & \textbf{VF1} \\
\midrule

\multirow{5}{*}{Gold}
& Gemma 2B       & 0.8393 & 0.4107 & 0.5824 \\
& Gemma 7B       & \textbf{1.0000} & 0.5778 & 0.8479 \\
& Mistral--7B    & 0.9942 & 0.5836 & 0.8408 \\
& Qwen2.5--7B    & 0.9950 & \textbf{0.6866} & \textbf{0.8930} \\
& LLaMA~3--8B    & 0.6542 & 0.3869 & 0.5510 \\
\midrule

\multirow{5}{*}{PaddleOCR}
& Gemma 2B       & 0.8624 & 0.3847 & 0.5735 \\
& Gemma 7B       & \textbf{0.9971} & 0.5173 & 0.8067 \\
& Mistral--7B    & 0.9870 & 0.4834 & 0.7824 \\
& Qwen2.5--7B    & 0.9935 & \textbf{0.5742} & \textbf{0.8367} \\
& LLaMA~3--8B    & 0.6837 & 0.3444 & 0.5500 \\
\midrule

\multirow{5}{*}{EasyOCR}
& Gemma 2B       & 0.7976 & 0.1988 & 0.3351 \\
& Gemma 7B       & 0.9856 & 0.3084 & 0.5621 \\
& Mistral--7B    & 0.9777 & 0.3372 & 0.6161 \\
& Qwen2.5--7B    & \textbf{0.9870} & \textbf{0.4236} & \textbf{0.6568} \\
& LLaMA~3--8B    & 0.9157 & 0.3501 & 0.5977 \\
\midrule

\multirow{5}{*}{Tesseract}
& Gemma 2B       & 0.8235 & 0.2363 & 0.3991 \\
& Gemma 7B       & \textbf{0.9510} & 0.3646 & 0.6410 \\
& Mistral--7B    & 0.9503 & 0.3653 & 0.6373 \\
& Qwen2.5--7B    & 0.9388 & \textbf{0.4330} & \textbf{0.6742} \\
& LLaMA~3--8B    & 0.7514 & 0.2961 & 0.5104 \\
\bottomrule
\end{tabular}
\end{table}

Gemma~7B obtains complete zero-shot KR, while Qwen2.5--7B obtains the highest EM and VF1. The difference between near-perfect key recovery and substantially lower EM shows that recognizing SROIE's compact field inventory is considerably easier than reproducing the corresponding values exactly.

\subsubsection{Few-Shot Prompting}

LLaMA~3--8B is particularly sensitive to demonstrations on SROIE: KR increases from 0.6542 at zero-shot to 0.9978 after one demonstration, with corresponding large gains in EM and VF1. Qwen2.5--7B begins with near-complete KR, so its few-shot gains are smaller and concentrated primarily in EM.

\subsubsection{OCR-Derived Inputs}

SROIE key recovery remains comparatively robust under OCR
degradation, whereas EM and VF1 decline more substantially.
For example, Table~\ref{tab:sroie-zero-shot} shows that
Qwen2.5--7B KR changes only from 0.9950 on Gold text to
0.9935 under PaddleOCR, while EM decreases from 0.6866 to
0.5742 and VF1 from 0.8930 to 0.8367. This pattern indicates
that OCR corruption often preserves enough information to
identify the field type while altering the value itself.

\subsubsection{Few-Shot Prompting under PaddleOCR}

Figure~\ref{fig:few_shot_trajectories}(d) shows the
corresponding SROIE PaddleOCR trajectories; exact scores are
reported in Table~\ref{tab:combined_fewshot}.


Few-shot prompting under PaddleOCR produces substantial gains for LLaMA~3--8B and more modest gains for Qwen2.5--7B. For Qwen2.5--7B, EM increases from 0.5742 at zero shot to 0.6715 at three shots, while VF1 increases from 0.8367 to 0.8561. LLaMA~3--8B shows a substantially larger response to demonstrations: KR increases from 0.6837 at zero shot to 0.9885 after one demonstration and reaches 0.9978 at three shots. Its highest EM is obtained at two shots, while its highest VF1 is obtained at three shots. As in the other experiments, additional demonstrations do not produce strictly monotonic improvements across all metrics.

\subsection{CORD Results}
\label{sec:results-cord}
CORD contains dense receipt information, repeated monetary values, and item-level fields. These characteristics create a different extraction challenge from both FUNSD and the compact SROIE schema.

\begin{table}[t]
\centering
\footnotesize
\setlength{\tabcolsep}{3pt}
\renewcommand{\arraystretch}{0.96}
\caption{Zero-shot KVP extraction performance on \textbf{CORD} across
Gold and OCR-derived text. Bold values indicate the best score
within each text condition.}
\label{tab:cord-zero-shot}
\begin{tabular}{@{}llccc@{}}
\toprule
\textbf{Text} & \textbf{Model} &
\textbf{KR} & \textbf{EM} & \textbf{VF1} \\
\midrule

\multirow{5}{*}{Gold}
& Gemma 2B       & 0.3140 & 0.2673 & 0.2682 \\
& Gemma 7B       & 0.7022 & 0.6121 & 0.6121 \\
& Mistral--7B    & 0.7820 & 0.6327 & 0.6361 \\
& Qwen2.5--7B    & \textbf{0.9045} & \textbf{0.8736} & \textbf{0.8748} \\
& LLaMA~3--8B    & 0.8581 & 0.7383 & 0.7383 \\
\midrule

\multirow{5}{*}{PaddleOCR}
& Gemma 2B       & 0.2179 & 0.1385 & 0.1385 \\
& Gemma 7B       & 0.5810 & 0.3486 & 0.3571 \\
& Mistral--7B    & 0.5885 & 0.3357 & 0.3401 \\
& Qwen2.5--7B    & \textbf{0.6927} & \textbf{0.4855} & \textbf{0.4906} \\
& LLaMA~3--8B    & 0.6879 & 0.4109 & 0.4207 \\
\midrule

\multirow{5}{*}{EasyOCR}
& Gemma 2B       & 0.1310 & 0.0576 & 0.0584 \\
& Gemma 7B       & 0.4477 & 0.1737 & 0.1809 \\
& Mistral--7B    & 0.4863 & 0.2087 & 0.2104 \\
& Qwen2.5--7B    & 0.5176 & \textbf{0.2698} & \textbf{0.2773} \\
& LLaMA~3--8B    & \textbf{0.5206} & 0.1888 & 0.1967 \\
\midrule

\multirow{5}{*}{Tesseract}
& Gemma 2B       & 0.1225 & 0.0413 & 0.0413 \\
& Gemma 7B       & 0.1972 & 0.0704 & 0.0734 \\
& Mistral--7B    & \textbf{0.2917} & 0.0857 & 0.0884 \\
& Qwen2.5--7B    & 0.2733 & \textbf{0.1469} & \textbf{0.1488} \\
& LLaMA~3--8B    & 0.2777 & 0.0919 & 0.0998 \\
\bottomrule
\end{tabular}
\end{table}

Table~\ref{tab:cord-zero-shot} shows that Qwen2.5--7B
obtains the highest zero-shot Gold-text scores, followed by
LLaMA~3--8B. Compared with FUNSD, KR and EM are much closer
for the strongest models, indicating that when CORD fields
are recovered under clean text, their values can often also
be extracted exactly.

\subsubsection{Few-Shot Prompting}

Figure~\ref{fig:few_shot_trajectories}(e) shows the CORD
Gold-text trajectories; exact scores are reported in
Table~\ref{tab:combined_fewshot}.

Both models initially benefit from demonstrations, but additional examples do not produce monotonic gains. Qwen2.5--7B reaches its highest scores at one shot and then declines, while LLaMA~3--8B reaches its highest scores at two shots. This behavior is consistent with the FUNSD experiments in showing that demonstration count alone is insufficient to characterize few-shot behavior.

\subsubsection{OCR-Derived Inputs}


As shown in Table~\ref{tab:cord-zero-shot}, CORD exhibits
substantial OCR sensitivity. For Qwen2.5--7B, VF1 decreases from 0.8748 on Gold text to 0.4906 under PaddleOCR, 0.2773 under EasyOCR, and 0.1488 under Tesseract. The deterioration is especially severe under Tesseract, consistent with its high CORD CER and WER. Model ordering is again metric-dependent: under EasyOCR, LLaMA~3--8B has slightly higher KR than Qwen2.5--7B, whereas Qwen2.5--7B retains higher EM and VF1.

\subsubsection{Few-Shot Prompting under PaddleOCR}


Figure~\ref{fig:few_shot_trajectories}(f) shows the CORD
PaddleOCR trajectories; exact scores are reported in
Table~\ref{tab:combined_fewshot}.

Under CORD PaddleOCR, Qwen2.5--7B reaches its highest scores with one demonstration and then declines, whereas LLaMA~3--8B improves across the evaluated shot counts. The contrast provides another example of model-specific responses to in-context demonstrations.

\subsection{Cross-Dataset Comparison}
\label{sec:results-cross}
The three datasets expose different failure regimes. FUNSD remains difficult even with corrected Gold text because flattening a form into a text sequence removes direct spatial association cues. SROIE, with its compact field inventory, permits near-complete key recovery for several models, although exact value extraction remains more sensitive to OCR corruption. CORD combines dense receipt content with repeated numeric values and exhibits particularly large degradation as OCR quality decreases.

The results also show that OCR robustness cannot be summarized by a universal ordering of models. A model that performs best under Gold text may not retain the same relative position under OCR-derived input, and differences depend on whether KR, EM, or VF1 is considered. Similarly, few-shot prompting is not uniformly beneficial: the magnitude and direction of the effect depend on the model, dataset, number of demonstrations, and, as the FUNSD sensitivity experiment shows, the composition of those demonstrations.

Across the three datasets, three OCR engines, and three evaluation metrics, the top-ranked zero-shot model under OCR-derived input differs from the corresponding Gold-text leader in 10 of 27 comparisons.

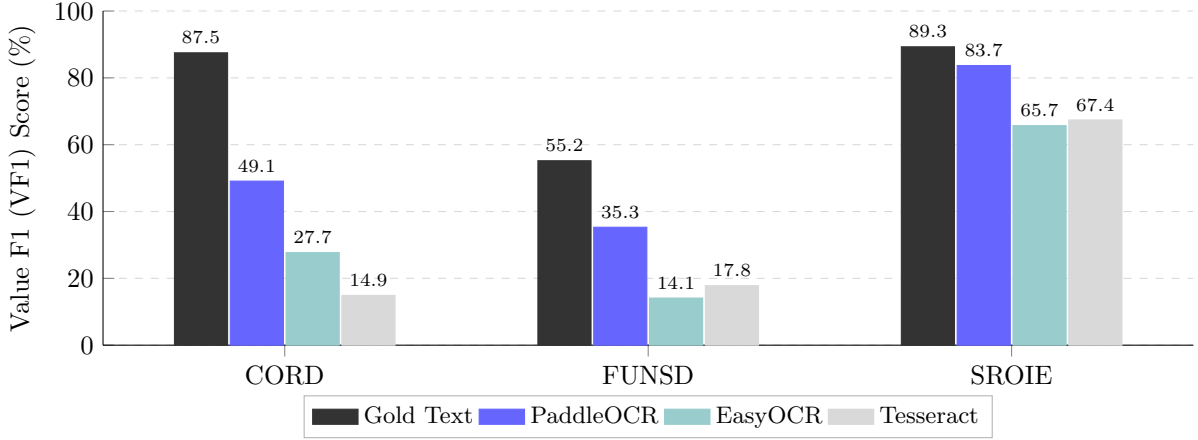
\begin{figure*}[htbp]
    \centering
    \begin{tikzpicture}
        \begin{axis}[
            ybar=1pt,
            bar width=20pt,
            width=\textwidth,
            height=6cm,
            enlarge x limits=0.25,
            area legend,
            legend style={
                at={(0.5, -0.15)}, 
                anchor=north,
                legend columns=-1, 
                draw=black!30,
                fill=white,
                font=\small
            },
            ylabel={Value F1 (VF1) Score (\%)},
            symbolic x coords={CORD, FUNSD, SROIE},
            xtick=data,
            ymin=0, 
            ymax=100,      
            ytick={0,20,40,60,80,100}, 
            clip=false, 
            nodes near coords,
            nodes near coords align={vertical},
            every node near coord/.append style={font=\footnotesize},
            ymajorgrids=true,
            grid style={dashed, gray!30},
            axis x line*=bottom,
            axis y line*=left,
            cycle list={
                {fill=black!80, draw=black!80},
                {fill=blue!60, draw=blue!60},
                {fill=teal!40, draw=teal!40},
                {fill=gray!30, draw=gray!30} 
            }
        ]
        
        \addplot coordinates {(CORD, 87.5) (FUNSD, 55.2) (SROIE, 89.3)};
        
        \addplot coordinates {(CORD, 49.1) (FUNSD, 35.3) (SROIE, 83.7)};
        
        \addplot coordinates {(CORD, 27.7) (FUNSD, 14.1) (SROIE, 65.7)};
        
        \addplot coordinates {(CORD, 14.9) (FUNSD, 17.8) (SROIE, 67.4)};
        
        \legend{Gold Text, PaddleOCR, EasyOCR, Tesseract}
        \end{axis}
    \end{tikzpicture}
    \caption{Zero-shot Value F1 for Qwen2.5--7B across Gold and
    OCR-derived inputs. The magnitude of degradation varies
    substantially across datasets and OCR engines.}
    \label{fig:vf1_degradation}
\end{figure*}

Figure~\ref{fig:vf1_degradation} visualizes this
dataset-dependent OCR degradation for Qwen2.5--7B.

\subsection{Observed Error Patterns}
\label{sec:results-qualitative}
During inspection of saved model outputs and evaluation artifacts, we observed several recurring error patterns across datasets and text conditions. This inspection was descriptive rather than a separately sampled or quantitatively coded qualitative study. Observed patterns included key--value misalignment, OCR-induced numeric corruption, fragmented values, unsupported or unannotated fields, and structured-output failures.

Key--value misalignment is particularly relevant on FUNSD, where flattening two-dimensional forms removes direct spatial association cues. On SROIE and CORD, OCR corruption of dates, addresses, and monetary values can preserve the general field identity while preventing exact value recovery, consistent with the larger OCR-induced reductions in EM and VF1 than in KR for several configurations. Repeated monetary values in receipt documents also create ambiguity when multiple predictions share the same normalized key.

Additional generated fields require careful interpretation because the quantitative evaluation is reference-centered. Some additional fields may represent unsupported generations, whereas others may correspond to semantically meaningful document content that is not included in a dataset's annotation schema. Accordingly, KR, EM, and VF1 are not interpreted as direct measurements of hallucination frequency.

\subsection{Comparison with Prior Document Understanding Approaches}
End-to-end VLMs and layout-aware models are not used as numerical baselines because they consume pixels, layout coordinates, or both and often employ task-specific training and entity-level metrics. Our protocol instead varies only the text representation supplied to fixed text-only LLM inference. Combining their published scores with our document-level KR, EM, and VF1 would conflate modality, supervision, target representation, and evaluation. Their results are therefore complementary and require a separate multimodal protocol.

\section{Discussion}
\label{sec:discussion}

The experiments show that text-only KVP extraction depends jointly on document structure, upstream text quality, model behavior, and prompting. The following sections discuss the implications of these interactions for OCR-based document understanding.

\subsection{Semantic Capability, OCR Quality, and Document Structure}

Across many model--dataset combinations, OCR affects exact value recovery more
strongly than field identification. For example, models may continue to
recognize that a receipt contains a \texttt{TOTAL} or \texttt{DATE} field while
OCR corruption changes one or more characters in the associated value. Such an
output can preserve key recovery while reducing Exact Match and Value~F1.
This distinction is important because apparently robust field detection does
not necessarily imply reliable extraction of the information required by a
downstream application.

The relationship between OCR quality and extraction performance is nevertheless
not completely uniform. Lower CER and WER generally correspond to stronger
downstream results at the dataset level, but the same aggregate OCR error rate
can conceal errors of very different semantic importance. Corrupting a
character in a monetary amount, date, or identifier may be more consequential
for KVP extraction than several errors in unrelated document text. OCR
robustness should therefore be understood in terms of both the amount and the
location of transcription error.

Document structure further mediates this relationship. FUNSD relies heavily on
spatial associations between questions and answers. Although spatially
reconstructing its annotation text provides a substantially more appropriate
clean-text representation than annotation-array ordering, the final LLM input
is still linear text and does not explicitly encode two-dimensional
relationships. SROIE, by contrast, has a compact field inventory and relatively
regular receipt structure, allowing high key recovery even when values are
partially corrupted. CORD combines dense receipt content, repeated numeric
values, and item-level information and exhibits particularly severe
degradation under poorer OCR conditions. These differences demonstrate why
robustness conclusions based on a single document dataset may not transfer
directly to other document types.

\subsection{Model Behavior under OCR Degradation}

Model behavior is not uniform across datasets, metrics, and text conditions.
Clean-text advantages do not necessarily persist after OCR degradation, and
the relative ordering of models can change with both OCR condition and
evaluation metric.

This behavior cautions against treating clean-text model rankings as direct
predictors of OCR robustness. OCR changes the evidence supplied to the model,
and different models respond differently to character corruption, altered
tokenization, fragmented values, and disrupted reading order. Consequently,
model selection for document extraction should be evaluated using textual
inputs representative of the intended deployment pipeline rather than inferred
solely from clean-text performance.

The results also do not support a simple claim that differences among models
always converge as OCR quality deteriorates. Severe OCR corruption can constrain
all models, as observed particularly on CORD under Tesseract, but relative
behavior remains dataset- and metric-dependent. OCR quality therefore acts as
an important constraint on downstream performance without eliminating
differences among model families.
\subsection{Few-Shot Prompting and Demonstration Sensitivity}

Few-shot effects are strongly model-dependent and non-monotonic, showing
that demonstration count alone does not determine extraction quality.

The confirmatory FUNSD sensitivity experiment further shows that demonstration
\emph{composition}, not only demonstration count, contributes to this
variability. Holding the number of demonstrations fixed at three while changing
their composition produces measurable differences in KR, EM, and VF1 for both
Qwen2.5--7B and LLaMA~3--8B. Low-complexity demonstration sets yield lower
mean VF1 than the representative strategy for both models, while increasing
complexity further does not produce monotonic improvement. Variation also
remains among independently sampled sets within the same strategy.

These results have two methodological implications. First, conclusions about
few-shot performance based on a single demonstration set can partly reflect
the particular examples selected rather than a general property of the model.
Reporting the exact demonstrations is therefore important for reproducibility.
Second, demonstration selection represents a meaningful experimental variable
in document extraction. More demonstrations, or more complex demonstrations,
are not automatically preferable; their usefulness depends on the model and
the extraction setting.

The context-feasibility audit further shows that demonstration
selection affects not only extraction performance but also
whether the complete target document can be accommodated under
a fixed context budget.

\subsection{Implications for Text-Only Document Understanding}

The benchmark highlights a fundamental limitation of text-only document
understanding. Once a document is converted into a linear text sequence, some
spatial, visual, and hierarchical information is no longer explicitly
available to the language model. Better prompting or a stronger language model
may improve interpretation of the information that remains, but cannot
reconstruct arbitrary evidence that has been removed or corrupted upstream.

This limitation is especially relevant to forms, tables, and visually complex
receipts in which semantic relationships depend on relative position rather
than text alone. At the same time, the results show that text-only LLMs can be
effective when the textual representation is sufficiently informative,
particularly because they enable flexible extraction without task-specific
fine-tuning.

Text-only LLMs and layout-aware or multimodal document models should therefore
be viewed as complementary design choices rather than directly interchangeable
approaches. Layout-aware architectures preserve structural information, while
instruction-tuned LLMs provide flexible semantic interpretation of textual
content. Systems that combine reliable OCR, structural representations, and
LLM-based reasoning may be better suited to heterogeneous real-world
documents, particularly when layout and visual context carry information that
cannot be preserved reliably in a flattened transcript.

\subsection{Implications for Benchmarking and Deployment}

The benchmark has three practical implications. First, annotation-derived text isolates downstream extraction behavior, whereas OCR-based evaluation remains necessary for deployment because aggregate CER and WER can conceal errors in task-critical values. Second, few-shot studies should report their demonstrations and assess sensitivity to their selection. Third, OCR, representation, prompting, and semantic extraction should be evaluated jointly: stronger LLMs cannot recover information lost upstream, and improved OCR cannot restore structural relationships removed during text serialization. Robustness should therefore be measured on the complete pipeline under target operating conditions.
\section{Limitations}
\label{sec:limitations}

The controlled benchmark has several limitations.

\subsection{Dataset and Evaluation Scope}

The evaluated datasets cover forms and receipts from a
limited set of document domains and languages. Although
many target fields are represented using English labels,
the source documents may contain non-English text.
The benchmark was not designed to systematically evaluate
cross-lingual generalization and does not cover the broader
diversity of handwritten, multi-page, or structurally complex
documents. Generalization beyond these settings therefore
remains to be established.

Evaluation is also constrained by dataset annotation coverage. SROIE, for
example, contains only four annotated fields (\textit{company}, \textit{date},
\textit{address}, and \textit{total}), although receipts may contain additional
semantically valid information. Predictions for unannotated fields are therefore
excluded rather than counted as false positives. Consequently, Key Recall,
Exact Match, and Value~F1 measure recovery of annotated key--value pairs rather
than precision over the complete set of generated fields. Unsupported or additional predictions were observed during
output inspection but were not quantified by the primary
evaluation metrics.

Finally, the study relies on automatic evaluation. Formatting differences,
annotation omissions, or semantically valid predictions outside the reference
schema may not be fully reflected by the reported scores; human evaluation
could provide complementary evidence of practical extraction quality.

\subsection{Experimental Design}

Results under OCR-derived inputs combine OCR and downstream extraction errors: substitutions, fragmentation, numeric corruption, reordering, and structural loss can all affect scores. Deterministic JSON-first parsing without semantic repair adds another failure source because malformed outputs may yield incomplete recovery despite partially correct content. We use a common prompt and decoding configuration for controlled comparison rather than per-model optimization.

Constrained- and open-schema prompting were not separately evaluated. SROIE lists its four fixed annotated fields, whereas FUNSD and CORD use open extraction because their keys vary by document. Schema listing may reduce unsupported outputs but suppress unlisted valid fields; open extraction offers greater flexibility but may generate additional fields. The reference-centered metrics do not quantify this precision--coverage trade-off.

\subsection{Model Coverage}
Resource constraints limited evaluation to instruction-tuned open-weight models of approximately 2B--8B parameters in a single-GPU environment. Whether the observed rank changes and non-monotonic few-shot behavior persist for substantially larger open-weight models or proprietary APIs remains open. Greater scale may improve instruction following and noise tolerance but cannot restore evidence absent from the serialized input. Task-specific fine-tuning, schema-guided or retrieval-augmented methods, and multimodal models are also outside scope because they introduce additional supervision or modalities.

\section{Conclusion}
\label{sec:conclusion}

This study presented a controlled benchmark of five open-weight instruction-tuned LLMs for text-only KVP extraction across FUNSD, SROIE, and CORD under annotation-derived and three OCR-derived text conditions. With prompting, inference, parsing, and evaluation fixed, OCR degradation affected exact and partial value recovery more strongly than field identification, and clean-text rankings did not reliably transfer to OCR-derived inputs. Few-shot prompting sometimes produced large gains but remained model-dependent, non-monotonic, and sensitive to demonstration composition.

These findings characterize text-only KVP extraction as a pipeline-level problem jointly governed by textual fidelity, document structure, model behavior, and demonstration choice. The released benchmark supports extension to additional domains, languages, OCR-aware methods, larger models, and multimodal systems.



\appendix
\section{Implementation Details}
\label{app:implementation}

All checkpoints used greedy decoding (\texttt{do\_sample=False}), an
8,192-token input limit, up to 1,024 generated tokens, batch size one, and no
task-specific fine-tuning. Runs were sequential on an NVIDIA A100 40\,GB using
Python 3.13.15, PyTorch 2.11.0, CUDA 12.8, and Transformers 5.16.1. OCR text
remained uncorrected, and JSONL artifacts supported resumable execution.

\begingroup
\setlength{\bibsep}{2pt}
\bibliography{refs}
\endgroup

\end{document}